%% file: root.tex
	\newif\ifboxes
	\documentclass[letterpaper, 10 pt, conference]{ieeeconf}  

	\IEEEoverridecommandlockouts                              
	
\newpage
	\title{\LARGE \bf
	Behavior-Tree-Based Person Search for Symbiotic \mbox{Autonomous Mobile Robot Tasks}
	}

	\author{Marvin Stuede, Timo Lerche, Martin Alexander Petersen and Svenja Spindeldreier$^{1}$
	\thanks{$^{1}$All authors are with the Leibniz University Hannover, Institute of Mechatronic Systems, D-30823 Garbsen, Germany,
	        {\tt\small marvin.stuede@imes.uni-hannover.de}}%
	}

	\newcommand{\capsize}{\fontsize{\small}\selectfont}
	\usepackage[compact]{titlesec}
	\usepackage[nolist]{acronym}
	\usepackage{cite}
	\usepackage{pdfpages}
	\usepackage{adjustbox}
	\usepackage{framed}
	\usepackage[ruled]{algorithm2e}
	\usepackage{nicefrac}
	\usepackage{amsmath}
	
	\DeclareMathOperator*{\argmin}{arg\,min}
	\usepackage{amssymb}
	\usepackage{subcaption}
	\usepackage[font=small]{caption}
	\usepackage[T1]{fontenc}
	\usepackage[utf8]{inputenc}
	\usepackage{grffile}
	\usepackage{siunitx}
	\usepackage{adjustbox}
	\usepackage{setspace}
	\usepackage{dblfloatfix} 
	\usepackage{pgfplots} 
	\pgfplotsset{compat=1.3,
		tick label style={font=\small,/pgf/number format/use comma}, 
		every axis legend/.append style={at={(0.5,1.03)},anchor=south,nodes=right,font=\small}, 
		label style={font=\capsize} 
	}
	\newlength\picwidth
	\newlength\boxwidth
	\newlength\tikzwidth
	\newlength\figureheight
	\newlength\figurewidth
	\newlength\shorten
		\newif\ifcopyright
	\copyrighttrue
	
\begin{document}

		\ifcopyright
					{\LARGE IEEE Copyright Notice}
		\newline
		\fboxrule=0.4pt \fboxsep=3pt
		
		\fbox{\begin{minipage}{1.1\linewidth}  
				Copyright (c) 2021 IEEE. Personal use of this material is permitted. For any other purposes, permission must be obtained from the IEEE by emailing pubs-permissions@ieee.org. \\
				
				Accepted to be published in: Proceedings of the 2021 IEEE International Conference on Robotics and Automation (ICRA), May 30 - June 5, Xi'an, China  
				
		\end{minipage}}
		\else
		\fi

	\begin{acronym}
		\acro{bt}[BT]{Behavior Tree}
		\acro{psbt}[PSBT]{\textit{Person Search Behavior Tree}}
		\acro{nw}[NW]{\textit{never wait}}
		\acro{w}[W]{\textit{wait at the help location}}
		\acro{gc}[GC]{\textit{greedy planning to a close maximum}}
		\acro{gm}[GM]{\textit{greedy planning to the global maximum}}
		\acro{rnd}[RND]{\textit{Uniform random sampling of goals}}
		\acro{sbt}[SBT]{Stochastic Behavior Tree}
		\acro{bts}[\textit{S}]{\textit{success}}
		\acro{btr}[\textit{R}]{\textit{running}}
		\acro{otsp}[OTSP]{open traveling salesman problem}
		\acro{btf}[\textit{F}]{\textit{failure}}
		\acro{dtmc}[DTMC]{Discrete Time Markov Chain}
		\acro{mdp}[MDP]{Markov Decision Process}
	\end{acronym}
	\newcommand*\prs[1][]{p_{\mathrm{s}#1}}
	\newcommand*\prf[1][]{p_{\mathrm{f}#1}}
	\newcommand{\wait}[1]{\mathcal{W}_{\mathrm{A},#1}}
	\newcommand{\place}[1]{\mathcal{P}_{#1}}
	\newcommand{\path}[2]{\mathcal{S}_{\mathrm{A},#1\to #2}}
	\newcommand{\waith}{\mathcal{W}_{\mathrm{A},\mathrm{h}}}
	\newcommand{\eal}{\textit{et al.}}
	\newcommand{\mvel}{\bar{v}}
	\newcommand{\lfail}{l_{\mathrm{fail}}}
	\newcommand{\ealcite}[1]{\eal~\cite{#1}}
	\newcommand{\bts}{\acs{bts}}
	\newcommand{\btr}{\acs{btr}}
	\newcommand{\btf}{\acs{btf}}
	\newcommand{\TODO}[1]{\textcolor{red}{\textbf{TODO: #1}}}
	\renewcommand{\vec}[1]{\mbox{\boldmath{$#1$}}}
	\newcommand{\dvec}[1]{\dot{\mbox{\boldmath{$#1$}}}}
	\newcommand{\ddvec}[1]{\ddot{\mbox{\boldmath{$#1$}}}}
	
	\newcommand{\transpose}{\ind{T}}
	
	\newcommand{\tmat}[2]{{^{\ind{#1}}\vec{T}_{\ind{#2}}}}
	\newcommand{\rmat}[2]{{^{\ind{#1}}\vec{R}_{\ind{#2}}}}
	\newcommand{\ks}[1]{{\ind{(KS)_{#1}}}}
	\maketitle
	\thispagestyle{empty}
	\pagestyle{empty}

	\begin{abstract}
	We consider the problem of people search by a mobile social robot in case of a situation that cannot be solved by the robot alone.
	Examples are physically opening a closed door or operating an elevator.
	Based on the Behavior Tree framework, we create a modular and easily extendable  action sequence with the goal of finding a person to assist the robot.
	By decomposing the Behavior Tree as a Discrete Time Markov Chain, we obtain an estimate of the probability and rate of success of the options for action, especially where the robot should wait or search for people.
	In a real-world experiment, the presented method is compared with other common approaches in a total of 588 test runs over the course of one week, starting at two different locations in a university building.
	We show our method to be superior to other approaches in terms of success rate and duration until a finding person and returning to the start location.
	\end{abstract}

	\section{INTRODUCTION}
\begin{figure}[t]
	\centering
	\includegraphics[width=1\linewidth]{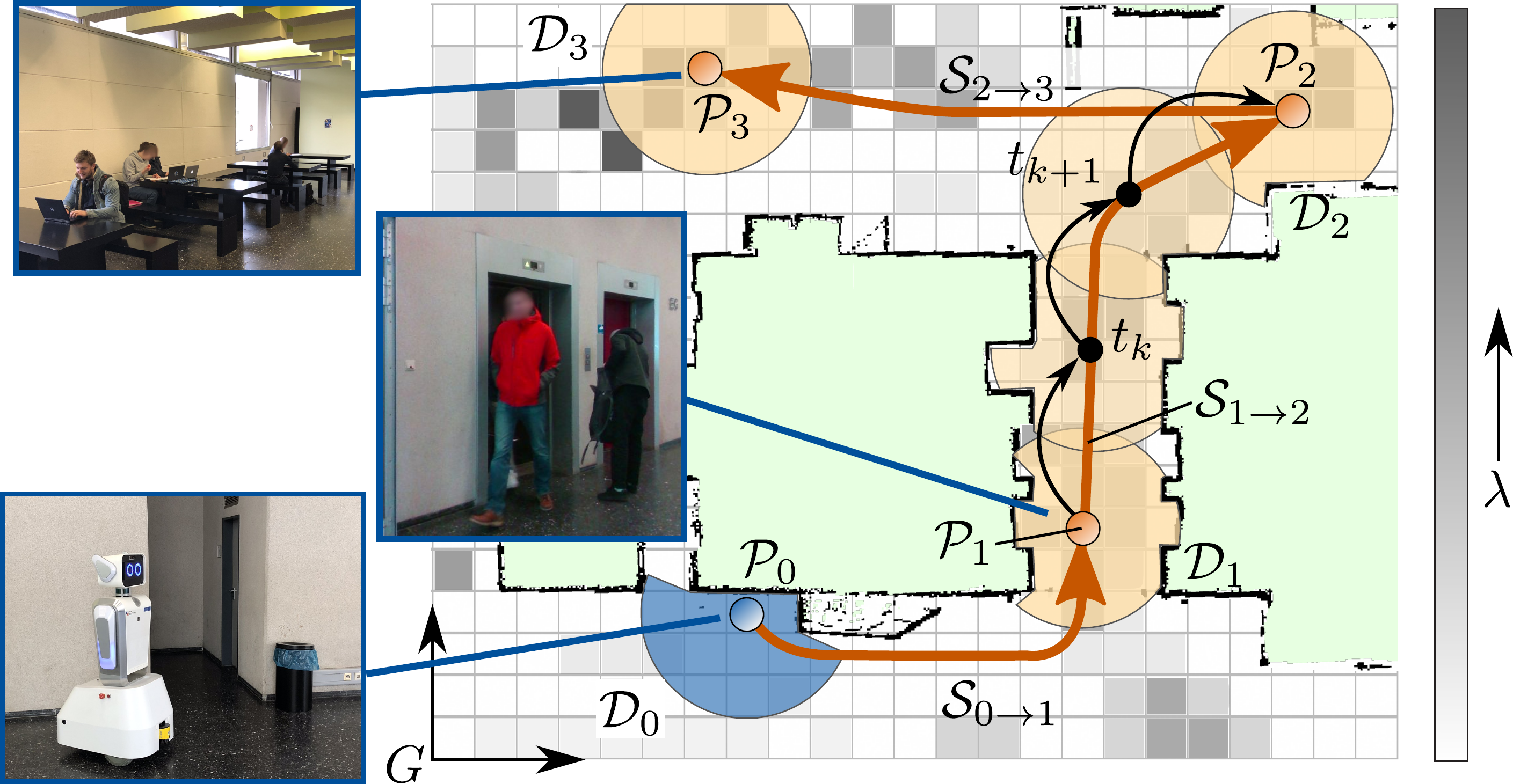}
	\caption{Exemplary arrangement of different search locations and paths.  $\mathcal{P}_{i}$ denotes a cell in the people occurrence model with rate $\lambda$, $\mathcal{D}_{i}$ a detection zone and $\mathcal{S}_{i\to j}$ a search path. For planning, the detection zone is slid along the search paths with discrete times $t_k$.}
	\label{fig:grid_schema}
	\vspace{-0.43cm}
\end{figure}

	Social service robots are entering an increasing number of areas of everyday life.
	Thanks to powerful interaction interfaces, often based on natural language, they are already being used commercially, for example to provide information \cite{pandey2018mass}, guiding or automated delivery \cite{ivanov2017adoption}.
	Contrary to their strength in socially interacting with people, they usually do not have physical manipulators for reasons of cost and complexity.
	This prevents them from making full use of the environment and, for example, opening doors \cite{stuede2019door} or operating elevators without further equipment \cite{wang2018robot}.
	An approach to compensate for these weaknesses is \textit{symbiotic autonomy}, which considers the recognition of an individually unsolvable situation and the active involvement of people in problem solving \cite{veloso2015cobots}.
	Depending on the problem, there are often no people available on site to help, so that helpers must be actively searched for.
	In this paper we utilize the Behavior Tree framework to find people in an open space based on a spatial model of people occurrence rate  (see Fig.~\ref{fig:grid_schema}).
	The method balances between a proactive search and waiting on site to avoid unnecessary travel and waiting times.
	 
	Behavior Trees \acused{bt}(\acp{bt}) are a procedural control approach that has become increasingly popular in the robotics community in recent years.
	Compared to other approaches, such as hierarchical finite state machines, they have clear advantages in terms of modularity, reusability or expandability \cite{Collendanchise14,paxton2017costar,colledanchise2018behavior}.
	In addition, they can be analyzed stochastically, which enables a prediction of the probability of success or failure of a particular tree while maintaining the interpretability of the tree's graphical representation \cite{Collendanchise14}.
	With \acp{bt}, in our approach the search for people is modelled flexibly and intuitively. Due to the modularity of the framework, the problem is defined at the level of tasks, such as \emph{wait at a location} or \emph{search along a path}. This avoids the usage of tailored low-level cost functions as used by comparable approaches  \cite{cas2014hmm,rosenthal2012is,tipaldi2011want}.
	The \acp{bt} are synthesized based on a stochastic model, which indicates the occurrence of people using spatially defined Poisson processes.
	We use the exponentially distributed inter-arrival times of these models as input for a decomposition of the \acp{bt} as \acp{dtmc} for the stochastic analysis.
	This work explicitly focuses on the search for people and excludes social aspects such as the appropriate approach to a person and verbal questions for help.
	However, the resulting trees can directly be integrated into a tree that models these aspects without further modification.
%
%
	\section{Related work}
	Overcoming the limitations of a robotic system by actively involving humans has been considered in various contexts. 
	Malfunctions can be mitigated by forewarning the users \cite{lee2010}  or purposefully utilizing human collaboration in autonomous plans \cite{nardi2014}.
	When a situation occurs where the robot needs help, this must first be identified, for example by detecting a closed door \cite{stuede2019door} or an elevator \cite{liebner2019door}.
	
	Independent of the search for help, different approaches to finding people were introduced, e.g. based on greedy search \cite{volkhardt2013finding}, hidden Markov Models \cite{cas2014hmm} or Periodic Gaussian Mixture Models \cite{krajnik2015s}.
	The search for symbiotic autonomous tasks has the additional requirement that people should not only be found, but then accompany the robot to the location where assistance is needed.
	Since people are only willing to travel a limited distance to help \cite{rosenthal2012mobile}, this imposes an additional constraint on the search locations.
	Most work in this context assumes that help is always available at the immediate help location, e.g. by supervisors \cite{tellex2014asking} or bystanders \cite{weiss2010robots}, and there is only little work on proactively searching and finding people to fulfill a task that cannot be achieved alone by the robot.
	Rosenthal \eal~\cite{rosenthal2012is} showed that navigating the environment to search for humans could decrease the time until a potential helper is found.
	They employ an A*-based planner to decide where to seek help in an office building based on the location of offices and availability of the person.
	However, their method is only evaluated with static locations, using occupancy sensors installed in offices and not applicable to dynamically created locations based on people detections.
	
	Approaches to model the occurrence of people generally use a spatial and temporal partitioning to assign a probability density to each region.
	Occurrences can then be modeled e.g. temporally via spectral analysis \cite{Krajnik2017} or spatially via direction identification  \cite{Senanayake2018}.
	Another common approach is to use a Poisson process \cite{jovan2016poisson} to stochastically model the number and time of occurrences of random events in a time interval.
	Ihler \eal~\cite{ihler2007learning} present a non-parametric Bayesian model of intensity functions representing events over time by learning the rate of a Poisson process for a spatially fixed scenario.
	The authors of \cite{tipaldi2011want} extend this approach by spatial and temporal variation of the rate parameter as a piecewise homogeneous Poisson process and then use this model to actively search for people.
	Although the approach is similar to ours in its theoretical basis, the problem is solved with an individually defined Markov Decision Process (MDP), which is only applicable with lattice-like movement primitives, and difficult to extend and integrate into more advanced models of symbiotic task execution.
	
	In summary, the application of dynamically created people models to symbiotic autonomy and a generally applicable action description to find people for this kind of tasks are open problems. The contributions of this paper are therefore: 1) A task descriptive model of a symbiotic autonomous person search, based on the Behavior Tree framework. 2) The connection of a stochastic environmental model with the \ac{bt} framework, complying with the requirements of a Stochastic Behavior Tree.  3) Real-world experiments showing the effectiveness of the approach, i.e. in reducing waiting times for spatially problematic helping tasks.
	
	The remainder of this paper is structured as follows: The next section \ref{sec:prem} gives a short overview over \acp{bt} and the stochastic people occurrence model.
	We then introduce how the stochastic actions are derived from the model in Sec.~\ref{sec:bt}.
	Sec.~\ref{sec:exp} experimentally shows the effectiveness of the approach compared to other methods and Sec.~\ref{sec:conc} gives a conclusion.
	
	\section{PRELIMINARIES}
	\label{sec:prem}

	\subsection{Behavior Trees}
	A \acf{bt} is a directed rooted tree, consisting of internal nodes for control flow and leaf nodes for action execution or condition evaluation \cite{colledanchise2018behavior}.
	Pairs of adjacent nodes are denoted as \textit{parent} (outgoing) and \textit{child} (incoming).
	The node without parents is called the root node, which periodically sends an enabling signal (\textit{tick}) through the tree that is propagated according to the policies of different control flow nodes.
	When a node receives a tick, it returns one of three status: \acl{btr}~(\btr), \acl{bts}~(\bts) or \ac{btf}.
	There are two types of leaf nodes: the action node, which returns \acs{bts} if an action was completed successfully and the condition node, which returns  \acs{bts} according to a pre-defined condition \cite{colledanchise2018behavior}.
	
	
	
	

	
%
%
	
	In the primary model of \acp{bt} the execution order of selector and sequence node children is inherently fixed and must be decided beforehand by the designer, which often is non-trivial.
	To overcome this problem, \acp{sbt} were introduced by Colledanchise \eal~\cite{Collendanchise14}.
	In an \ac{sbt} each action $\mathcal{A}$ is extended by a tuple $\mathcal{A}_{\mathrm{sbt}}:\,\left(\prs(t), \prf(t), \mu, \nu \right)$ and each condition $\mathcal{C}$ by a tuple $\mathcal{C}_{\mathrm{sbt}}:\left(\prs(t), \prf(t)\right)$, where $\prs(t)$ $(\prf(t))$ is the probability to succeed (fail) at any given time $t$.
	The time to succeed (fail) is a random variable with exponential distribution and rate $\mu$ $(\nu)$.
	By describing the inner flow of an \ac{sbt} as a \acf{dtmc}, an indication of the success probability $\prs[,\mathrm{T}](t)$ of the whole tree can be made by summing up the probabilities of being in one of the \ac{dtmc} success states $\mathcal{S}_{\mathrm{S}}$:
	\begin{equation}
	\label{eq:suc_tree}
	\prs[,\mathrm{T}](t)=\sum_{i: s_{i} \in \mathcal{S}_{\mathrm{S}}} \pi_{i}(t).
	\end{equation}
	The probability vector $\pi(t)$ is obtained by solving the Cauchy problem 
	\begin{equation}
	\label{eq:cauchy}
	\dot{\pi}(t) = Q(t) \pi(t),\quad \pi(0) = \pi_0,
	\end{equation}
	with $Q$ as the infinitesimal generator matrix of the \ac{dtmc}.
	The success rate $\mu_{\mathrm{T}}$ of the tree  is calculated as
	\begin{equation}
	\label{eq:rate_tree}
	\mu_{\mathrm{T}}=\mathrm{avg}\left(\frac{\sum_{i=1}^{\left|\mathcal{S}_{\mathrm{S}}\right|} u_{i 1}^{\mathrm{S}}(\kappa) \log \left(h_{i 1}^{\mathrm{S}}(\kappa)\right)}{\sum_{i=1}^{\left|\mathcal{S}_{\mathrm{S}}\right|} u_{i 1}^{\mathrm{S}}(\kappa)}\right)^{-1}
	\end{equation}
	with $\mathrm{avg}(.)$  as the average function over time and $\kappa$ as a time step.
	The matrices $H^\mathrm{S}(\kappa,\boldsymbol{\mathcal{A}}_{\mathrm{sbt}},\boldsymbol{\mathcal{C}}_{\mathrm{sbt}})$ and $U^\mathrm{S}(\kappa,\boldsymbol{\mathcal{A}}_{\mathrm{sbt}},\boldsymbol{\mathcal{C}}_{\mathrm{sbt}})$ depend on the transit times, number of steps between transient states and success states and the success and failure probabilities and rates of actions $\boldsymbol{\mathcal{A}}_{\mathrm{sbt}}$ and conditions $\boldsymbol{\mathcal{C}}_{\mathrm{sbt}}$.
	The sets $\boldsymbol{\mathcal{A}}_{\mathrm{sbt}}$ and $\boldsymbol{\mathcal{C}}_{\mathrm{sbt}}$ contain all individual actions and conditions of the tree.
	The failure probability of the tree $\prf[,\mathrm{T}](t)$ follows likewise based on the failure states $\mathcal{S}_{\mathrm{F}}$. For more in-depth explanations we refer to \cite{Collendanchise14} and \cite{colledanchise2018behavior}.
	
	\subsection{Probabilistic People Occurrence Model}
		\label{sec:prob}
	The common approach for probabilistic temporal and spatial description of the occurrence of events, e.g. the appearance of people within an area, is the Poisson process.
	In general, a Poisson process is a renewal process with Poisson-distributed random variable $\left(N(t),t\geq0\right)$.
	The probability of $N(t)$ being equal to a count $c$ is given by
	\begin{equation}
	P\left(N(t)=c\right)=\frac{(\lambda t)^{c}}{c !} e^{-\lambda t}\quad\mathrm{with}\quad c=0,1,2,...,
	\end{equation}
	where $\lambda$ is the rate parameter of the process.
	When $\lambda(t)$ is variable, the process is called \textit{inhomogeneous}.
	An inhomogeneous \textit{spatial} Poisson process introduces an additional spatial dependency $\vec{x} \in \mathbb{R}^d$ in an Euclidean space $\mathbb{R}^d$ for the rate $\lambda(\vec{x},t)$.
	As proposed in \cite{luber2011}, for $\vec{x} \in \mathbb{R}^2$ the inhomogeneous spatial Poisson process can be approximated by a 2D grid representation
	\begin{equation}
	\label{eq:grid}
	G\,:\quad\lambda(\vec{x}, t) \simeq \sum_{i=1}^m\sum_{j=1}^o \lambda_{i j \tau} \mathbf{1}_{i j \tau}(\vec{x})
	\end{equation}
	with $G: \mathbb{R}^{m \times o} \rightarrow \mathbb{R}$, indicator function $\mathbf{1}_{i j \tau}(\vec{x})$ and $\lambda_{i j \tau}$ as the constant rate of a piecewise homogeneous Poisson process, valid in a time interval $ [t_\tau,t_{\tau+1})$.
	
	Learning the probabilistic representation of people occurrences can then be achieved by learning the constant rates $\lambda_{i j \tau}$.
	For a confidence-sensitive estimation of the rate parameter, Bayesian inference is used with a Gamma-distributed prior $\lambda_{\tau} \sim \Gamma\left(\lambda_{ \tau};\alpha_{\tau},\beta_{\tau}\right)$ (indices $i,j$ omitted for brevity).
	The shape parameter $\alpha_{\tau}$ and inverse scale parameter $\beta_{\tau}$ are determined incrementally. In our case, for a discrete time step $\sigma$ for all $t_\sigma<t_{\tau}$  the update rules
	\begin{align}
	\alpha_\sigma = \alpha_{\sigma-1} + c_\sigma\,\mathbf{1}_\mathcal{D}(\vec{x}_{\mathrm{R}}, t_\sigma),\quad \beta_{\sigma} = \beta_{\sigma - 1} + \mathbf{1}_\mathcal{D}(\vec{x}_{\mathrm{R}}, t_\sigma)
	\end{align}
	with initial values $\alpha_0=\beta_0=1$ and the number of detected people $c_\sigma\in\mathbb{N}$ since the last time step are used.
	The indicator function $\mathbf{1}_\mathcal{D}(\vec{x}_{\mathrm{R}}, t_\sigma)$ results from the detection area $\mathcal{D}$ of the robot at pose $\vec{x}_{\mathrm{R}}\in\mathbb{R}^2$ and causes that only the grid cells which lie within the detection range of the robot will be updated.
	We use a 3D Lidar for person detection, which  provides a full 360-degree environmental view, and therefore approximate $\mathcal{D}$ by a circle with radius $r$.
	By providing an estimation of person encounter probability at a specific location, the people occurrence model forms the basis for the decision whether the robot should wait at the place where help is needed or actively search for help.


	\section{Behavior Tree based person search}
		\label{sec:bt}
	The goal of the \ac{bt} description is to set up a sequence of actions to maximize the probability of meeting a person, or in other words, to determine if and where the robot should search for or wait for people.
   To create the tree, the atomic actions $\wait{i}$: \textit{Wait at place $\mathcal{P}_{i}$} and $\path{i}{j}$: \textit{Search from place $\mathcal{P}_{i}$ to place  $\mathcal{P}_{j}$} are defined, where a \textit{place} $\mathcal{P}_{i} \in G$ refers to a specific cell in the people occurrence grid $G$ (eq.~\ref{eq:grid}).
	\subsection{Definition of Atomic Actions}
	\label{sec:bt_actions}
	For this section we assume that there is a number of $n$  known places the robot could move to and/or wait at, including the robot's position $\mathcal{P}_{0}$, which is the location where help is needed (Sec. \ref{sec:choosing_bt} shows how $n$ is determined).
	To decide between different behaviors (i.e. Behavior Trees), the tuple $\mathcal{A}_{\mathrm{sbt}}$ must first be defined for each type of action.
	The wait action $\wait{i}$ will return \bts~when a person is found and \btf~when a maximum time has been reached.
	The success rate $\mu_\mathrm{w}$ then results directly from the property that the time differences between events of the Poisson process (i.e. interarrival times of people) are exponentially distributed.
	When the robot starts waiting at a time $t_0$, the probability density function of the exponential distribution
	\begin{equation}
	f(t ; \mu_\mathrm{w})=\left\{\begin{array}{ll}
	{\mu_\mathrm{w} e^{-\mu_\mathrm{w} t}} & {t \geq t_0} \\
	{0} & {t<t_0}
	\end{array}\right.,\quad\mu_\mathrm{w}=\sum_{\mathcal{D}}\lambda_{i j \tau}
	\end{equation}
	 describes the waiting time.
	 This requires the assumption that $(t-t_0) < (t_{\tau+1} - t_{\tau})$ i.e., the rates $\lambda_{i j \tau}$ can be regarded as constant while waiting.
	The success rate $\mu_\mathrm{w}$ results from the accumulated rates of all visible cells at the waiting position.
	The probability to meet a person while waiting is given by the corresponding cumulative distribution function
	\begin{equation}
	\prs[,\mathrm{w}](t ;\mu_\mathrm{w})=\left\{\begin{array}{ll}
	{1-e^{-\mu_\mathrm{w} t}} & {t \geq t_0} \\
	{0} & {t<t_0}
	\end{array}\right..
	\end{equation}
	By manually specifying a desired confidence $ \prs^{\prime} > \prs[,\mathrm{w}](\mu_\mathrm{w}^{-1} ;\mu_\mathrm{w})$, this equation can be rearranged to estimate the expected waiting time $T^{\prime}$ until the next person appears.
	If no person appears after $T^{\prime}$, the wait action is considered as failed, resulting in the mean time to fail:
	\begin{equation}
	\label{eq:rate_fail_wait}
	\nu_\mathrm{w}^{-1}=T^{\prime}=-\log(1-\prs^{\prime})\,\mu_\mathrm{w}^{-1}.
	\end{equation}
	Because the wait action will never fail before $T^{\prime}$ has passed, the fail probability is defined as 
	\begin{equation}
	\prf[,\mathrm{w}](t ;\mu_\mathrm{w})=\left\{\begin{array}{ll}
		{1-\prs^{\prime}} & {t \geq t_0+T^{\prime}} \\
		{0} & {t<t_0+T^{\prime}}
		\end{array}\right..
	\end{equation}
	
	For proactive search, we define a search path as
	\begin{equation}
	\mathcal{S}_{i\to j}:\: \left(\mathcal{P}_{i}, \mathcal{P}_{j}, \mathcal{G}, l, \mvel \right),\quad i,j\in\lbrace0,1,...,n \rbrace,\,i\neq j,
	\end{equation}
	where $\mathcal{P}_{i}$ and $\mathcal{P}_{j}$ are the start and end places, $\mathcal{G}\subset\mathbb{R}^2$ is the geometric description, $l$ is the length of the path and $\mvel$ the average velocity of the robot while driving on the path.
	$\mathcal{G}$ and $l$ directly follow from a geometric path planner (such as the A* algorithm on an occupancy grid map) and $\mvel$ can either be determined empirically or based on the  settings of a local path planner.
	Similar to the waiting action, we define the searching action $\path{i}{j}=\mathcal{A}_\mathrm{sbt}\cup\mathcal{S}_{i\to j}$ to return \bts~as soon as a person is found and \btf~if the whole path was driven without finding anyone.
	Additionally, this action can also return \btf~if the navigation execution fails, e.g. due to an obstructed goal.
	While the robot moves on the path, it observes different cells of the people occurrence grid $G$, each for an individual time span.
	This imposes a time dependency on the success rate $\mu_{\mathrm{sp}}(t)=\sum_{\mathcal{D}(t)}\lambda_{i j \tau}$ ("$\mathrm{sp}$" denotes \emph{search path}).
	The rate is calculated by discretizing the path with  $t_k=k\,\Delta t$ and calculating the corresponding rate $\mu_{\mathrm{sp},k}$ for each point in time.
	Fig.~\ref{fig:grid_schema} illustrates this as an example of different paths in an environment.
	
	The minimal time until a person is found on the path results from the sum of the time $t_k$ and the expected arrival time (as in eq. \ref{eq:rate_fail_wait}) to
	\begin{equation}
	\label{eq:path_rate}
	\mu_\mathrm{sp,tot}^{-1}=\argmin_{t_k \in \left[t_{0},t_{0}+\nicefrac{l}{\mvel} \right]}\left(t_k-\log(1-\prs^{\prime})\,\mu_{\mathrm{sp},k}^{-1}\right).
	\end{equation}
	Here, $\prs^{\prime}$ again is a confidence value and $t_{0}$ the time when the search is started.
	The total success rate $\mu_\mathrm{sp,tot}$ of the search path is the inverse of this time.
	The success probability $\prs[,\mathrm{sp}](t)$ of finding a person on the path follows as the counter probability of not having found anyone until time $t$ to
	\begin{equation}
	\prs[,\mathrm{sp}](t;\mu_\mathrm{sp})=1-\exp{\int_{t_{0}}^{t}\mu_\mathrm{sp}(\tilde{t})\mathrm{d}\tilde{t}}.
	\end{equation}
	If nobody has been found until the goal is reached or the navigation fails, the action \emph{search person} is considered as failed.
	The fail rate $\nu_\mathrm{sp}$ is approximated by
	
	\begin{equation}
	\label{eq:sp_fail_rate}
	\nu_\mathrm{sp} = \frac{\mvel}{l} + \frac{\mvel}{l_{\mathrm{fail}}}, 
	\end{equation}
	where $l_{\mathrm{fail}}$ is an expected distance the robot can move until a navigation failure occurs, which we also assume as exponentially distributed.
	This value can for instance be estimated by observation.
	Before the proactive search action is finished, the action can only fail due to the navigation and only after it has been finished, due to it not finding a person.
	The probability $\prf[,\mathrm{sp}](t;\nu_\mathrm{sp})$ to fail is therefore defined as a piecewise function:
	\begin{equation}
	\label{eq:sp_fail_prob}
	\prf[,\mathrm{sp}](t;\nu_\mathrm{sp})=\begin{cases}
	1 - \prs[,\mathrm{sp}](\nu_\mathrm{sp}^{-1}),& t \geq  \nu_\mathrm{sp}^{-1}\\
	1 - \exp(-\frac{\mvel}{\lfail}\,t), & t<\nu_\mathrm{sp}^{-1} \\
	\end{cases}.
	\end{equation}
	Introducing $l_{\mathrm{fail}}$ into eq. \ref{eq:sp_fail_rate} thus increases the probability to fail for longer search paths, giving preference to shorter paths with otherwise equal chance of finding a person.
	For  eq. \ref{eq:sp_fail_prob} to be applicable, the condition
	\begin{equation}
	\prs[,\mathrm{sp}](\nu_\mathrm{sp}^{-1}) \leq\exp(-\frac{l}{\lfail+l})
	\end{equation}
	must hold due to the law of total probability.
	This is fulfilled for $l \ll \lfail$, which is the case for the present application, since all longer paths can be discarded to avoid searching far away from the help location.
	\subsection{Choosing a Behavior Tree}
	\label{sec:choosing_bt}
\begin{figure}[b]
		\vspace{-0.5cm}
		\centering
		
		\includegraphics[width=\linewidth]{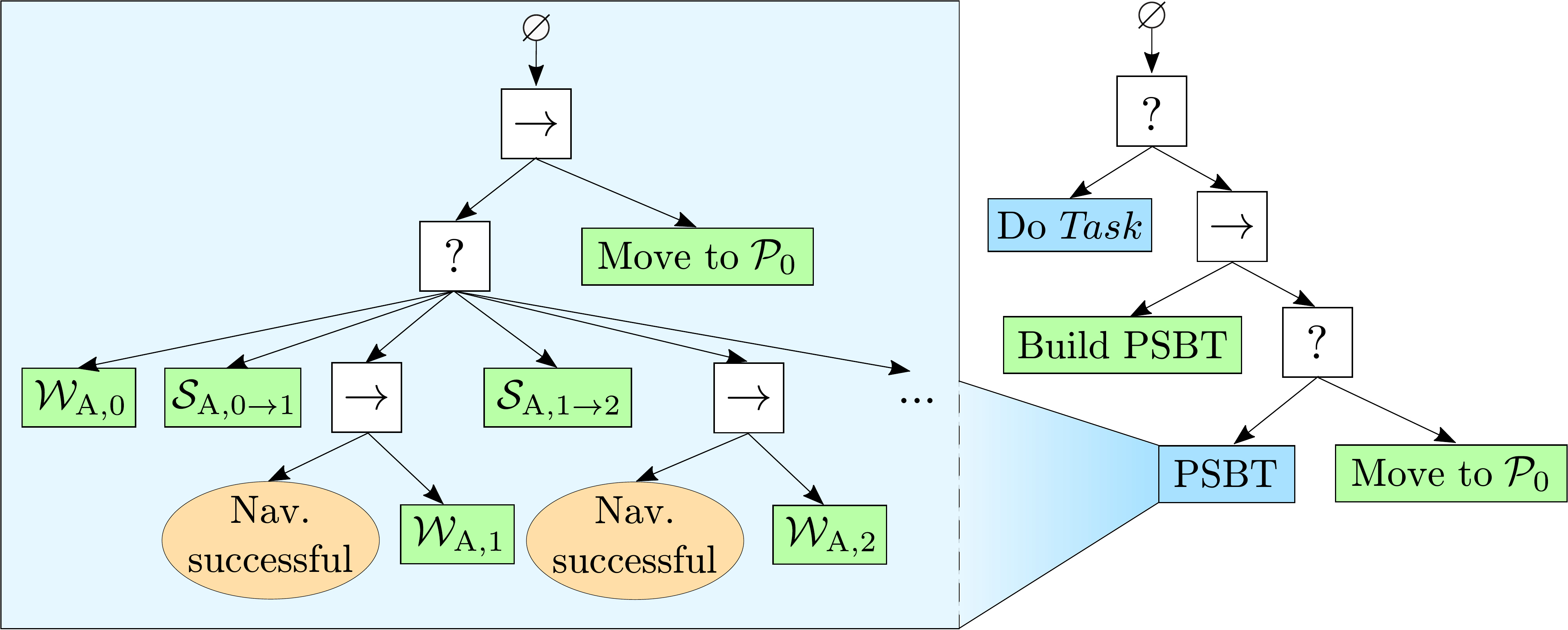}
		\caption{General form of the \ac{psbt}. This tree is executed as a fallback of a $Task$ where the robot needs help, e.g. opening a door or operating an elevator.}
		\label{fig:bt_general}
		\vspace{-0.25cm}
	\end{figure}

	Based on the wait and search actions, an action rule must now be found that maximizes the probability of finding a person, taking into account the return time to the help location.
	We therefore define the \ac{psbt}, which describes a sequence of actions that should be executed when the robot is facing a task that cannot be solved by itself.
	For this, a selector behavior is defined, with a general form as shown in Fig.~\ref{fig:bt_general}.
	The tree contains a $\mathrm{Move~to~}\mathcal{P}_{0}$ action, which describes the movement back to the start location and is interpreted as a search action that can only fail due to the navigation.

		\begin{figure}[b]
				\vspace{-0.5cm}
	\centering
	\begin{minipage}[t]{.49\linewidth}
		\includegraphics[width=\linewidth]{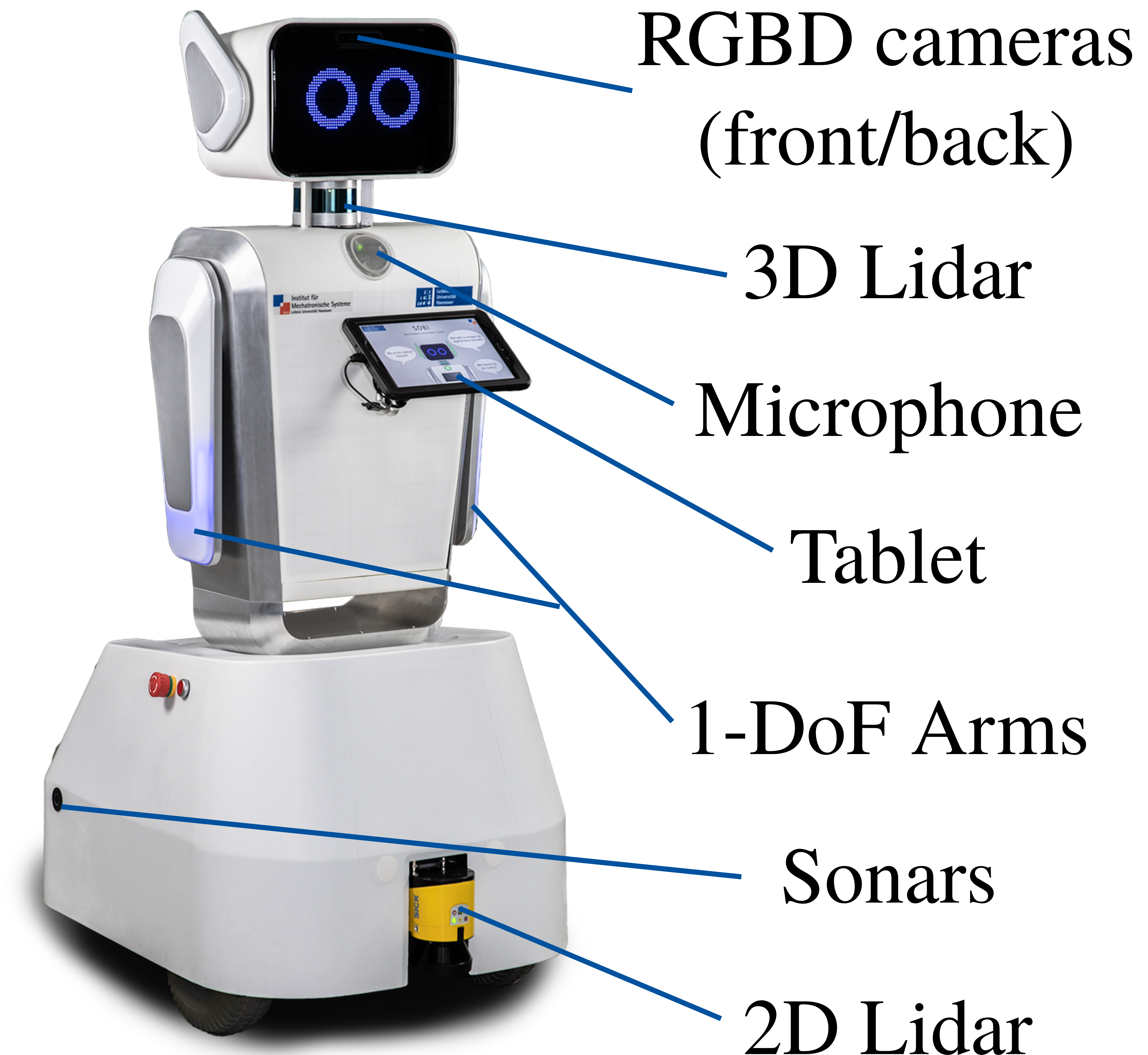}
		\captionof{figure}{\small{The ROS-based mobile social robot Sobi.}}
		\label{fig:sobi}	
	\end{minipage}
	\hfill
	\begin{minipage}[t]{.49\linewidth}
		
		\includegraphics[width=\linewidth]{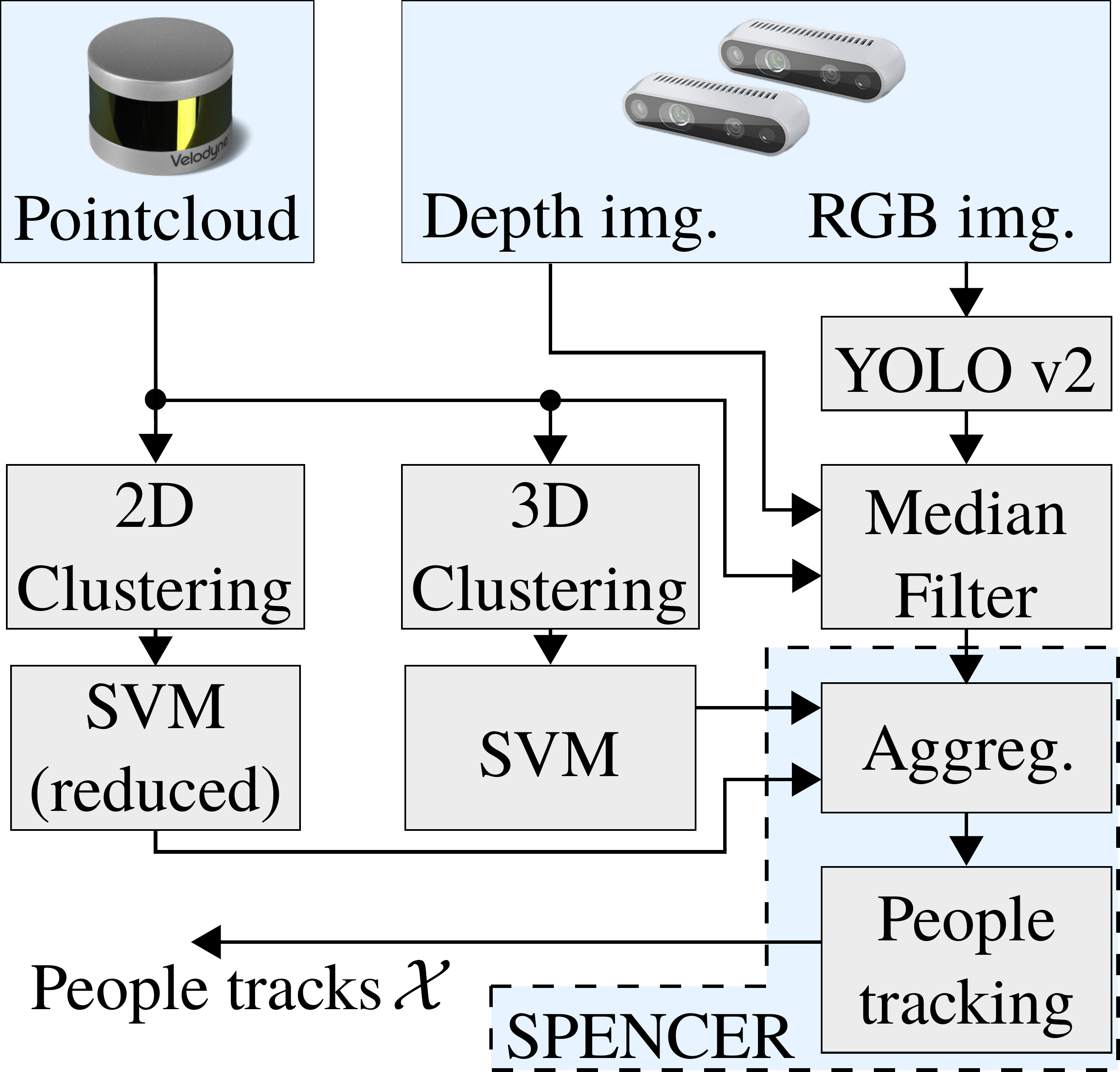}
		
		\captionof{figure}{Person perception and tracking pipeline.}
		\label{fig:percept}
	\end{minipage}

\end{figure}
	To create the \ac{psbt}, we first sample a number of $n$ places from the people occurrence grid $G$ via roulette wheel selection, where the probability $p$ of a cell to be selected is $p \propto \lambda_{i j}$.
	If sampled cells are close to each other (the Euclidean distance is smaller than the detection radius $r$), only the cell with the larger rate $\lambda_{i j}$ is kept.
	Also, only cells with a variance and distance to the robot below specified thresholds can be sampled.
	Subsequently, all wait actions $\wait{i}$ and search paths $\path{i}{j}$ ($\forall i,j\in\lbrace0,1,...,n \rbrace,\,i\neq j$) are calculated.
	To reduce the complexity $\mathcal O(n!)$ of investigating all possible sequences to search all places $\place{1...n}$, we determine the search order by means of an \ac{otsp} with places $\place{0...n}$ as nodes and the inverse of the failure rates $\nu_{\mathrm{s},i\to j}^{-1}$ as costs for the corresponding graph.
	The \ac{otsp} is solved by genetic algorithm \cite{davendra2010traveling}.
	The next step is the stochastic analysis of all possible \acp{bt}, with the goal to obtain an estimate of the success probability $\prs[,\mathrm{T}](t)$ and success rate $\mu_{\mathrm{T}}$ of the tree  (according to eq. \ref{eq:suc_tree} and eq. \ref{eq:rate_tree}) as the basis for decision-making.
	Each tree's flow is decomposed as a \ac{dtmc} by solving the Cauchy problem (eq. \ref{eq:cauchy}) with the infinitesimal generator matrix $Q(\boldsymbol{\mathcal{A}}_{\mathrm{sbt,sp}},\boldsymbol{\mathcal{A}}_{\mathrm{sbt,w}})$ until a pre-defined look-ahead time $t_{\mathrm{max}}$.
	We calculate this for every possible case, namely driving to specific places (or not),  waiting there (or not) and finally returning to the help location, leading to a total number of $3^n+1$ executions.
	The preferred tree is then chosen as the tree with maximum $\prs[,\mathrm{T}](t_{\mathrm{max}})$.
	
	%
	%
	%
	%
	%
	%
			\begin{figure*}[b]
	\vspace{-0.37cm}
	\begin{minipage}[t]{0.6\textwidth}
		
		\setlength\figureheight{3.5cm}
		\setlength\figurewidth{1\textwidth}
		
		\captionsetup{width=\linewidth}
		
		\trimbox{0.3cm 0.25cm 0.4cm 0.18cm}{
			\small{
				\input{fig/tikz/t150.tex}
			}
		}
		\caption{Probability distribution of the \ac{bt} root nodes $\prs[,\mathrm{T}](t_k)$ at seven points in time for 500 randomly sampled starting poses on the map.}
		\label{fig:model}
	\end{minipage}
	\hspace{1cm}
	\begin{minipage}[t]{0.3\textwidth}
		\setlength\figureheight{3.5cm}
		\setlength\figurewidth{1\textwidth}
		
		\captionsetup{width=\linewidth}
		\trimbox{0.55cm 0.32cm 0.4cm -0.1cm}{	
			\input{fig/tikz/t150_mtts.tex}			
		}
		\caption{Expected time to success $\mu_{\mathrm{T}}^{-1}$.}
		\label{fig:sim_mtts}
	\end{minipage}
	\vspace{-0.2cm}
\end{figure*}
	\section{Experiments}

		\label{sec:exp}
	The experimental evaluation is divided into two parts: The evaluation of the performance of the stochastic \ac{psbt} based on the people occurrence model and a comparison of the time to find people under real life conditions with the predicted time from the model.
	The environment for evaluation is the ground floor of a multi-storey \SI{50}{\meter} by \SI{25}{\meter} university building.
	The area includes lecture halls, a cafeteria, several entrances and sitting areas (see Fig.~\ref{fig:grid_schema}).
	For the real-world experiments, the mobile social robot Sobi (see Fig. \ref{fig:sobi}) is used, which is capable of people perception and tracking through 3D Lidar and RGBD cameras.
	We use an individually trained version of \cite{yan2017class} for 3D pointcloud segmentation, extended by the approach proposed in \cite{bogoslavskyi2016fast} for 2D clustering to decrease the false negative rate.
	In addition, YOLO v2 \cite{redmon2017yolo9000} is used for detection in the RGB images, which is then registered via a median filter based on the distance either with the 3D Lidar data or the depth data of the cameras.
	Aggregation and tracking is performed with the SPENCER framework \cite{linder2016multi}, which outputs people tracks $\mathcal{X}$ in Cartesian space with uncertainty information.
	Fig.~\ref{fig:percept} shows an overview of the perception and tracking pipeline.
	The model is trained with data from two working days by moving the robot throughout the day to different places on the entire floor, so that the same locations are observed at different times of day.
	Only people tracks with a minimum observed time of \SI{3}{\second} in a range of \SI{5}{\meter} are used for training, resulting in a total number of $18,362$ tracks $\mathcal{X}$.
	This rather high number resides in the fact that the robot partly loses the same person for a short time and then re-detects them as a new track.
	If a person moves more than one meter or is detected for more than \SI{20}{\second}, this is also considered as a new track.
	The following parameters were chosen for the experiments: $m=50$, $o=25$ (cell resolution of \SI{1}{\meter}), detection radius $r=\SI{2}{\meter}$,  confidence $\prs^{\prime}=0.9$, path discretization $\Delta t=\SI{1}{\second}$, expected distance $l_{\mathrm{fail}}=\SI{100}{\meter}$, average velocity $\mvel=\SI{0.5}{\meter\per\second}$, sampled goals $n=6$, \ac{dtmc} look-ahead time $t_{\mathrm{max}}=\SI{200}{\second}$.

	\subsection{\ac{psbt}: Model-based Evaluation}

	We compare the \ac{psbt} with five other strategies. The heuristic method of \ac{gm} plans to the cell with maximum occurrence rate $\lambda$ and waits there. The method \ac{gc} is similar to \ac{gm}, but plans to the closest cell out of the  $n_\lambda$ cells with highest rate and waits there ($n_\lambda=50$ in this evaluation).  \ac{rnd} samples the same number of cells $n$ like \ac{psbt} without regarding the rates of the cells and randomly waits at the locations.
	Furthermore, the trivial approach to \ac{w} and the strategy to sample the goals in the same manner as the \ac{psbt}, but \textit{never wait} at sampled locations to only search proactively (\acs{nw}) are implemented. 
	\newlength{\foursubht}
	\newsavebox{\foursubbox}

	First, the different methods are compared offline based on the trained people occurrence model.
	For this, we randomly sample 500 accessible poses on the occupancy grid map of the building as starting poses for the search for helpers.
	Then the probability of success $\prs[,\mathrm{T}](t_k)$ is calculated using the methods presented in section \ref{sec:bt_actions}.
	The comparison is shown in Fig.~\ref{fig:model} for seven points in time (every \SI{20}{\second}).
	Although the observable variance shows a dependency on the starting pose, \ac{psbt} outperforms the other methods, leading to an on average \SI{5.7}{\percent} higher  probability of success after \SI{140}{\second} than the next best method (\ac{gm}).
	The \ac{psbt} is always at least as good as waiting at the help location (\ac{w}), because the stochastic analysis also explicitly considers this case and can execute it accordingly.
	Fig.~\ref{fig:sim_mtts} shows the expected time to success $\mu_{\mathrm{T}}^{-1}$ for the four best methods.
	The time $\mu_{\mathrm{T}}^{-1}$ refers to the expected time to find a person and then return to the help location.
	Here, the expected time for the \ac{psbt} is averagely \SI{11.05}{\second} higher than for \ac{gm}, which is due to the property that the \ac{dtmc} decomposition yields more conservative estimates of the success rate for an increasing number of selector child nodes.
	Additionally, the resulting probability distribution $\prs[,\mathrm{T}](t_k)$ from the \ac{dtmc} is not exponential, therefore it is not exactly correlated to $\mu_{\mathrm{T}}^{-1}$.
	Fig.~\ref{fig:examples} shows two exemplary help locations in front of doors and the corresponding \acp{bt}.
	The location $\mathcal{P}_{\mathrm{stor}}$ (Figs.~ \ref{fig:path_1}, \ref{fig:tree_1}) is in front of a door to a storage room without a large number of people coming and going.
	While the greedy strategies aim for the closest (or global) maxima, the \ac{psbt} strategy moves to the nearby building entrance $\mathcal{P}_1$, waits there and then proceeds to move along the lifts to the global maximum (a sitting area).
	The second location is one of the entrances to the cafeteria $\mathcal{P}_{\mathrm{cafe}}$ (Figs.~ \ref{fig:path_2}, \ref{fig:tree_2}), which is located in an area with a higher volume of people.
	Here the \ac{psbt} strategy first waits on site before proactively searching for people, while the greedy strategies again aim for the maxima.

	\subsection{\ac{psbt}: Real-World Experiments}
		\begin{figure*}[t]
		\vspace{2mm}
		\setlength\figureheight{3cm}
		\sbox\foursubbox{%
			\resizebox{\dimexpr\textwidth-1em}{!}{%
				\includegraphics[height=\figureheight]{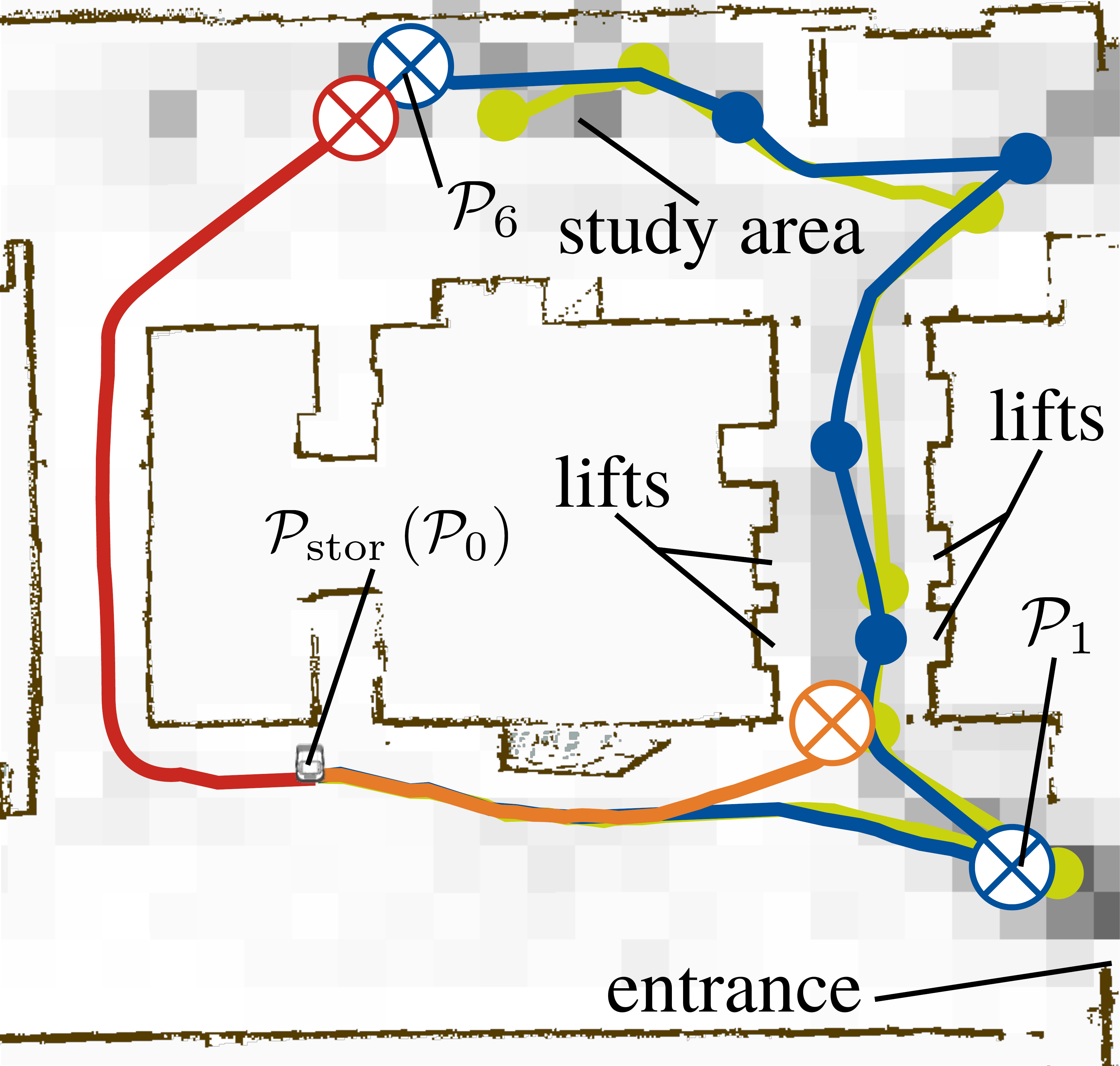}%
				\includegraphics[height=\figureheight]{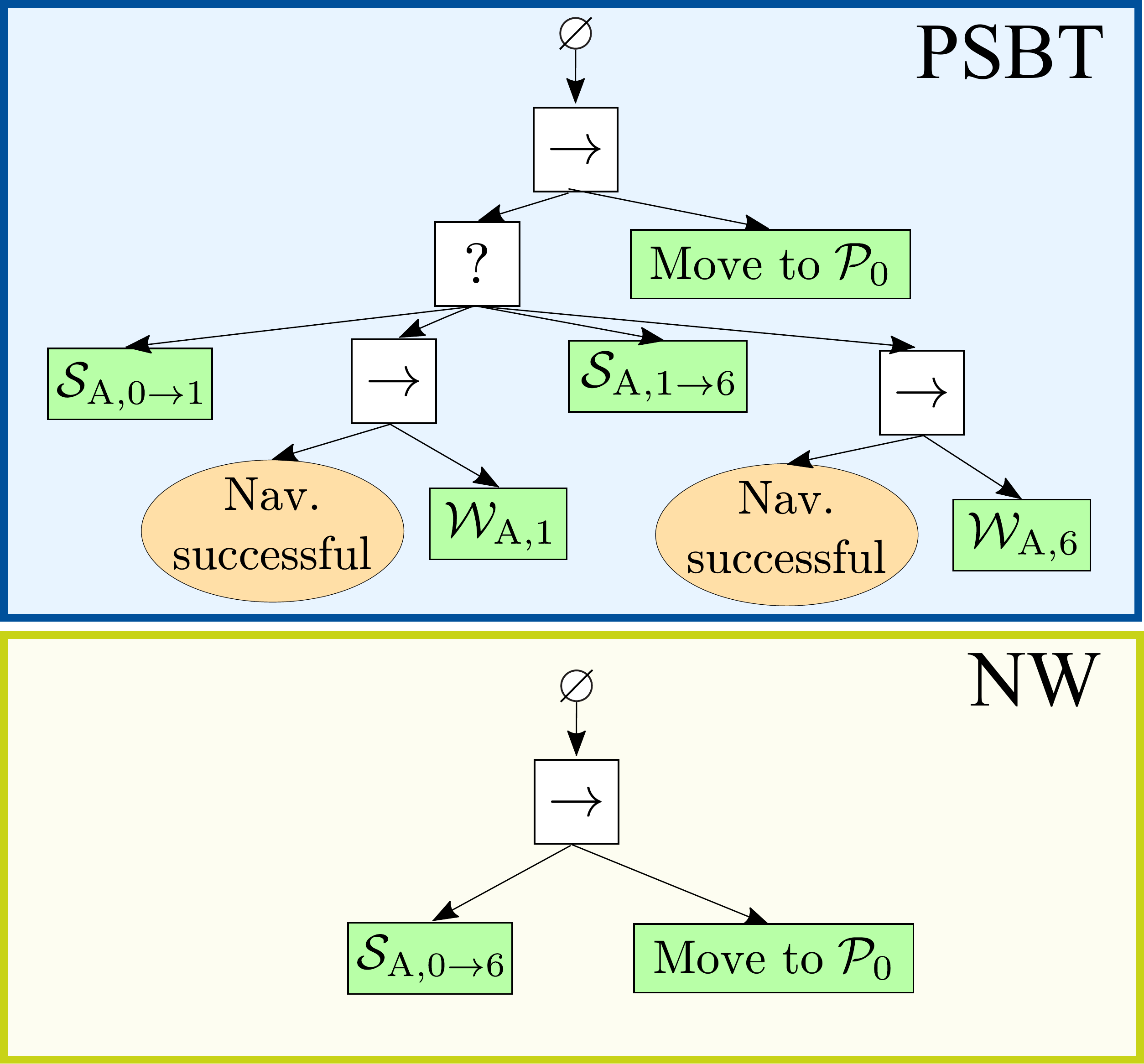}%
				\includegraphics[height=\figureheight]{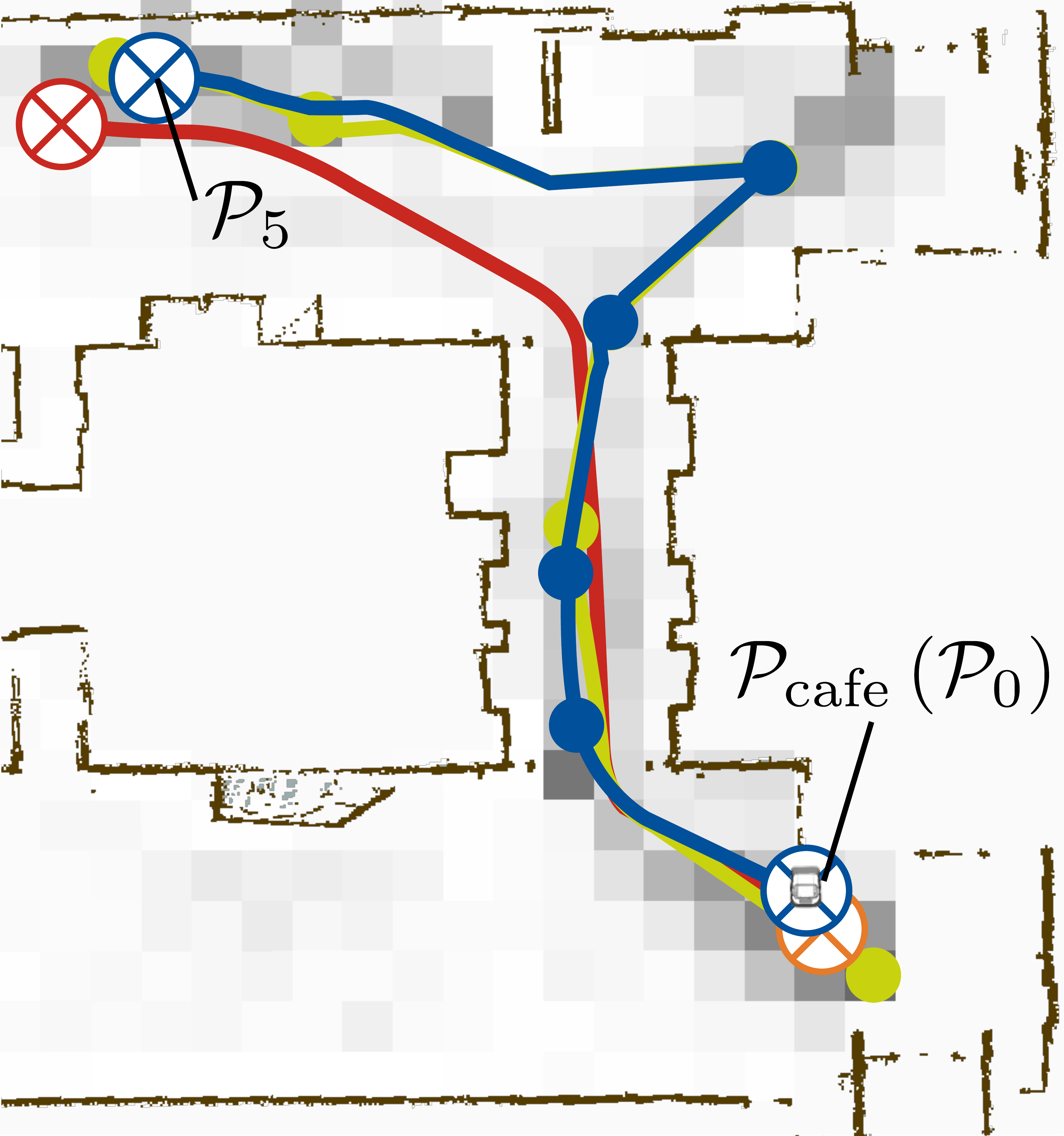}%
				\includegraphics[height=\figureheight]{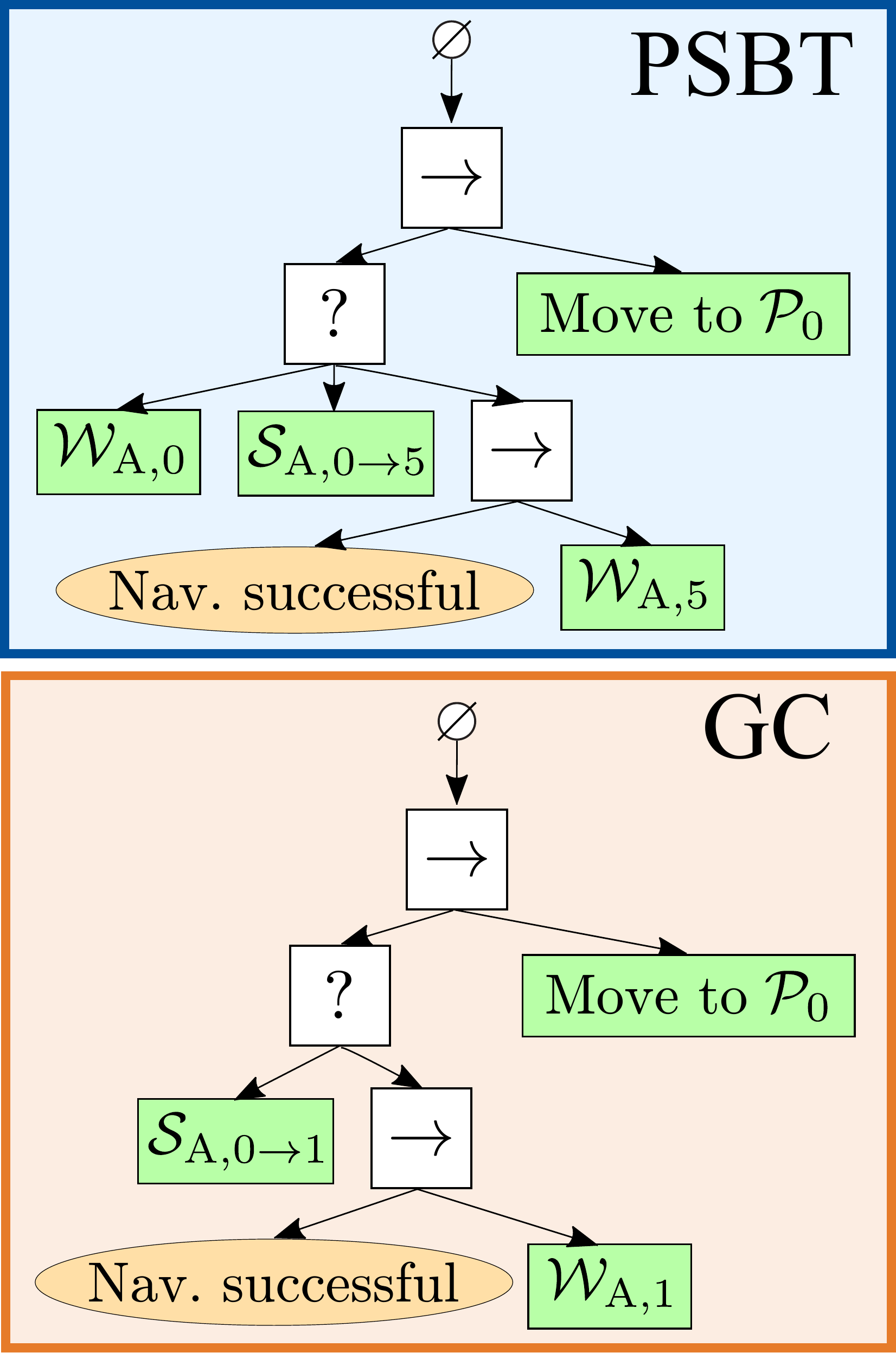}%
			}%
		}
		\setlength{\foursubht}{\ht\foursubbox}
		\centering
		
		\subcaptionbox{\label{fig:path_1}}{%
			
			\includegraphics[height=\foursubht]{fig/path_example.pdf}%
			
		}
		\subcaptionbox{\label{fig:tree_1}}{%
			\includegraphics[height=\foursubht]{fig/bt_example_sbt.pdf}%
		}
		\subcaptionbox{\label{fig:path_2}}{%
			\includegraphics[height=\foursubht]{fig/path_example_2.pdf}%
		}
		\subcaptionbox{\label{fig:tree_2}}{%
			\includegraphics[height=\foursubht]{fig/bt_example_2.pdf}%
		}
		\vspace{-2mm}
		\caption{Exemplary search paths planned by the different methods and corresponding search \acp{bt} for a help location $\mathcal{P}_{\mathrm{stor}}$ in an area with low rate $\lambda$, and a location $\mathcal{P}_{\mathrm{cafe}}$ with high rate. Colored dots indicate sampled places and crossed circles indicate a waiting location. Consecutive search paths are merged if there are no wait actions inbetween. Colors correspond to the legend in Fig.~\ref{fig:model}.}
		\label{fig:examples}
		
		\setlength\boxwidth{0.2\textwidth}
		\vspace{3mm}
		\begin{minipage}[b]{\boxwidth}
			\setlength\figureheight{2.7cm}
			\setlength\figurewidth{0.9\linewidth}
			\centering
			\trimbox{0.2cm 0.0cm 0.0cm 0.0cm}{
				\input{fig/tikz/02_13_sbt_drive.tex}
			}
			\captionsetup{font=small}
			\captionof{figure}{$t_{\mathrm{r}}$ for successful searches from $\mathcal{P}_{\mathrm{stor}}$}
			\label{fig:plot_comp_sto}
		\end{minipage}%
		\hfill
		\begin{minipage}[b]{\boxwidth}
			\vspace{0pt}
			\setlength\figureheight{2.7cm}
			\setlength\figurewidth{0.9\linewidth}
			\centering
			\trimbox{0.35cm 0.0cm 0.0cm 0.0cm}{
				\input{fig/tikz/02_14_sbt_drive.tex}
			}
			\captionsetup{font=small}
			\captionof{figure}{$t_{\mathrm{r}}$ for successful searches from  $\mathcal{P}_{\mathrm{cafe}}$}
			\label{fig:plot_comp_caf}
		\end{minipage}
		\hfill					
		\begin{minipage}[]{0.5\textwidth}
			\vspace{-43mm}
			\captionsetup{font=small}
			\captionof{table}{Results of the experiments in the university building with averagely estimated expected time to success $\bar{\mu}_\mathrm{r}^{-1}$ and averagely estimated success probability $\bar{p}_{\mathrm{s,T}}(t_{\mathrm{max}})$. Quantitative results given as mean $\pm$ std. deviation.}
			\begin{adjustbox}{width=\linewidth,center}
				
				\input{table/places.tex}
			\end{adjustbox}
			
			\label{tab:places}
		\end{minipage}
		\vspace{-6mm}	
	\end{figure*}
	To demonstrate the efficiency of the presented approach, we also conducted real-world experiments.
	For the two different start locations from Fig.~\ref{fig:examples}, we tested the different methods over a period of five working days.
	The same constant people occurrence model was used for all methods, which were tested in an alternating fashion in order to compensate for biasing effects (e.g. events in the lecture halls).
	A single run includes online-planning according to the corresponding method and the subsequent search for people.
	Instead of approaching the sampled goals directly, the closest reachable pose in a radius of \SI{2}{\meter} around the goal is chosen, allowing the sensors to be oriented towards the goal.
	We consider a person as found if a detected track $\mathcal{X}_i$ of a person is within the radius $r$ in front of the robot for a period of at least \SI{3}{\second}.
	After a successful search, the robot returns to the starting position and the time for the whole run $t_{\mathrm{r}}$ is saved.
	A run is considered as failed, when the search tree returns \btf , i.e. when all search and wait actions were executed without finding anyone.
	Successful searches based on false positive person detections are removed manually from the evaluation.
	
	The waiting time for all methods is determined according to eq.~\ref{eq:rate_fail_wait} with the confidence value $\prs^{\prime}$.
	Fig.~\ref{fig:plot_comp_sto} shows the results for $\mathcal{P}_{\mathrm{stor}}$ as the starting location.
	All proactive methods perform significantly better than waiting at the help location (\ac{w}).
	Table \ref{tab:places} summarizes the mean time $\bar{t}_{\mathrm{r}}$ with standard deviation, which were calculated for the successful runs only.
	The shortest average time $\bar{t}_{\mathrm{r}}$ was achieved by the methods \ac{gc} and \ac{gm}.
	Although the greedy methods can be faster, \ac{psbt} delivers substantially better results in terms of the success rate $P$. 
	Since the greedy methods only approach one maximum and remain there, the success depends strongly on whether people can be recognized at this specific location.
	For example, the global maximum in this environment was a resting area, often occupied by people in slightly different spots, posing larger difficulties for detection on a static waiting location.
	This is also reflected in the high success rate of the \acs{nw} method, as it is more likely to detect a person directly in front of the robot during the proactive search.
	A low success rate is also problematic in that it takes time to determine that all actions have failed (here always more than \SI{200}{\second}), therefore PSBT would still be faster on average compared to GC/GM.
	An essential advantage of the \ac{psbt} method can be seen for highly frequented areas like $\mathcal{P}_{\mathrm{cafe}}$ (see Fig.~\ref{fig:plot_comp_caf}).
	Here we compare the two best methods from  $\mathcal{P}_{\mathrm{stor}}$, to work out the difference between exclusively proactive search and intermittent waiting.
	The exclusively proactive \ac{nw} method takes longer on average to find a person and return to the start location, since \ac{psbt} correctly decides that it is more worthwhile to wait on site.
	Although not evaluated, it should be noted that the greedy strategy \ac{gc}  would lead to similar results as \ac{psbt} here (see e.g. the waiting location of \ac{gc} in Fig.   \ref{fig:path_2}).
	Based on the expected time to success $\mu_\mathrm{r}^{-1}$ and success probability $\prs[,\mathrm{T}](t_{\mathrm{max}})$ the model gives an estimate of the performance of the different methods and the necessary time to find a person, thus providing a good estimation for decision making.
	It provides more conservative estimates for the duration, which is due to the fact that cells with high variance are ignored in the calculation.
	Furthermore, the assumed average velocity of the robot $\mvel$ during the movement, was often exceeded in the test runs.
	The calculation time of the \ac{psbt} averaged \SI{4.8}{\second} for the test runs in this section (Intel i7-7700T CPU), which allows for online execution.
	To summarize, in 198 trial runs, our method was able to produce suitable predictions of the time until success and found people in \SI{94}{\percent} of all cases, which is a higher rate than all other compared methods.

	\section{CONCLUSION}
		\label{sec:conc}
		\addtolength{\textheight}{-7.8cm} 
	We presented a method for finding people in situations where a mobile robot needs human help to solve problems.
	Based on locally defined Poisson processes to model the occurrence rate of people in an environment, a Behavior Tree is created, linking consecutive search and wait actions to find a person in the shortest possible time.
	Compared to purely proactive search, the search time can be significantly reduced by actively deciding for or against waiting at the help location while maintaining the success rate.
	The BT framework enables the modular formulation of the search problem, as well as easy future expandability to include further actions, such as approaching and talking to people or accompanied drives to the help location.

	


	



	
	\bibliography{library}
	\bibliographystyle{IEEEtran}
	
	\end{document}

%% file: fig/tikz/t150.tex
%
%

\definecolor{mycolor1}{rgb}{0.00000,0.31373,0.60784}%
\definecolor{mycolor2}{rgb}{0.00000,0.44700,0.74100}%
\definecolor{mycolor3}{rgb}{0.90588,0.48235,0.16078}%
\definecolor{mycolor4}{rgb}{0.85000,0.32500,0.09800}%
\definecolor{mycolor5}{rgb}{0.92900,0.69400,0.12500}%
\definecolor{mycolor6}{rgb}{0.49400,0.18400,0.55600}%
\definecolor{mycolor7}{rgb}{0.46600,0.67400,0.18800}%
\definecolor{mycolor8}{rgb}{0.30100,0.74500,0.93300}%
\definecolor{mycolor9}{rgb}{0.78431,0.15686,0.12549}%
\definecolor{mycolor10}{rgb}{0.78431,0.82745,0.09020}%
\definecolor{mycolor11}{rgb}{0.90588,0.96863,1.00000}%
\begin{tikzpicture}

\begin{axis}[%
width=\figurewidth,
height=0.984\figureheight,
at={(0\figurewidth,0\figureheight)},
scale only axis,
xmin=0.5,
xmax=7.5,
xtick={1,2,3,4,5,6,7},
xticklabels={{$t_{20}$},{$t_{40}$},{$t_{60}$},{$t_{80}$},{$t_{100}$},{$t_{120}$},{$t_{140}$}},
ymin=0,
ymax=1,
ytick={0,0.2,0.4,0.6,0.8,1},
yticklabels={{$0$},{$0.2$},{$0.4$},{$0.6$},{$0.8$},{$1$}},
ylabel style={font=\color{white!15!black}},
ylabel={$\prs[,\mathrm{T}]$},
axis background/.style={fill=white},
ymajorgrids,
legend style={at={(0.03,0.97)}, anchor=north west, legend cell align=left, align=left, draw=white!15!black},
ylabel style={rotate=-90, at={(axis description cs:0.01,0.94)}, anchor=north east,inner ysep=0pt},legend columns=2,legend style={font=\scriptsize},major tick length=0
]
\addlegendimage{area legend,fill=mycolor1}
\addlegendimage{area legend,fill=mycolor3}
\addlegendimage{area legend,fill=mycolor9}
\addlegendimage{area legend,fill=mycolor10}
\addlegendimage{area legend,fill=mycolor11}
\addlegendimage{area legend,fill=white}
\addplot [color=black, dashed]
  table[row sep=crcr]{%
0.708333333333333	0.333023189141526\\
0.708333333333333	0.0636723749339581\\
};
\addplot [color=black]
  table[row sep=crcr]{%
0.685	0.333023189141526\\
0.731666666666667	0.333023189141526\\
};
\addplot [color=black]
  table[row sep=crcr]{%
0.685	0.0636723749339581\\
0.731666666666667	0.0636723749339581\\
};
\draw[fill=mycolor1, draw=black] (axis cs:0.661666666666667,0.103430908173323) rectangle (axis cs:0.755,0.195472395080172);
\addplot [color=black, line width=1.0pt]
  table[row sep=crcr]{%
0.661666666666667	0.138648220803589\\
0.755	0.138648220803589\\
};
\addplot [color=black, dashed]
  table[row sep=crcr]{%
1.70833333333333	0.574716808917726\\
1.70833333333333	0.125087385997176\\
};
\addplot [color=black]
  table[row sep=crcr]{%
1.685	0.574716808917726\\
1.73166666666667	0.574716808917726\\
};
\addplot [color=black]
  table[row sep=crcr]{%
1.685	0.125087385997176\\
1.73166666666667	0.125087385997176\\
};
\draw[fill=mycolor1, draw=black] (axis cs:1.66166666666667,0.211956841871142) rectangle (axis cs:1.755,0.360597517943378);
\addplot [color=black, line width=1.0pt]
  table[row sep=crcr]{%
1.66166666666667	0.262723042240995\\
1.755	0.262723042240995\\
};
\addplot [color=black, dashed]
  table[row sep=crcr]{%
2.70833333333333	0.7288274649544\\
2.70833333333333	0.191336341202259\\
};
\addplot [color=black]
  table[row sep=crcr]{%
2.685	0.7288274649544\\
2.73166666666667	0.7288274649544\\
};
\addplot [color=black]
  table[row sep=crcr]{%
2.685	0.191336341202259\\
2.73166666666667	0.191336341202259\\
};
\draw[fill=mycolor1, draw=black] (axis cs:2.66166666666667,0.304830491542816) rectangle (axis cs:2.755,0.503828344917517);
\addplot [color=black, line width=1.0pt]
  table[row sep=crcr]{%
2.66166666666667	0.367816045880318\\
2.755	0.367816045880318\\
};
\addplot [color=black, dashed]
  table[row sep=crcr]{%
3.70833333333333	0.827092757708284\\
3.70833333333333	0.259174033999443\\
};
\addplot [color=black]
  table[row sep=crcr]{%
3.685	0.827092757708284\\
3.73166666666667	0.827092757708284\\
};
\addplot [color=black]
  table[row sep=crcr]{%
3.685	0.259174033999443\\
3.73166666666667	0.259174033999443\\
};
\draw[fill=mycolor1, draw=black] (axis cs:3.66166666666667,0.390308186411858) rectangle (axis cs:3.755,0.610347514534514);
\addplot [color=black, line width=1.0pt]
  table[row sep=crcr]{%
3.66166666666667	0.454582244157791\\
3.755	0.454582244157791\\
};
\addplot [color=black, dashed]
  table[row sep=crcr]{%
4.70833333333333	0.889749474695515\\
4.70833333333333	0.324415422976017\\
};
\addplot [color=black]
  table[row sep=crcr]{%
4.685	0.889749474695515\\
4.73166666666667	0.889749474695515\\
};
\addplot [color=black]
  table[row sep=crcr]{%
4.685	0.324415422976017\\
4.73166666666667	0.324415422976017\\
};
\draw[fill=mycolor1, draw=black] (axis cs:4.66166666666667,0.463076800107956) rectangle (axis cs:4.755,0.693998925827009);
\addplot [color=black, line width=1.0pt]
  table[row sep=crcr]{%
4.66166666666667	0.533688064664602\\
4.755	0.533688064664602\\
};
\addplot [color=black, dashed]
  table[row sep=crcr]{%
5.70833333333333	0.929701161334772\\
5.70833333333333	0.385284587740898\\
};
\addplot [color=black]
  table[row sep=crcr]{%
5.685	0.929701161334772\\
5.73166666666667	0.929701161334772\\
};
\addplot [color=black]
  table[row sep=crcr]{%
5.685	0.385284587740898\\
5.73166666666667	0.385284587740898\\
};
\draw[fill=mycolor1, draw=black] (axis cs:5.66166666666667,0.530235391664064) rectangle (axis cs:5.755,0.759691877024307);
\addplot [color=black, line width=1.0pt]
  table[row sep=crcr]{%
5.66166666666667	0.599010360898077\\
5.755	0.599010360898077\\
};
\addplot [color=black, dashed]
  table[row sep=crcr]{%
6.70833333333333	0.955175481440733\\
6.70833333333333	0.440690220998817\\
};
\addplot [color=black]
  table[row sep=crcr]{%
6.685	0.955175481440733\\
6.73166666666667	0.955175481440733\\
};
\addplot [color=black]
  table[row sep=crcr]{%
6.685	0.440690220998817\\
6.73166666666667	0.440690220998817\\
};
\draw[fill=mycolor1, draw=black] (axis cs:6.66166666666667,0.589968574267914) rectangle (axis cs:6.755,0.811281727934543);
\addplot [color=black, line width=1.0pt]
  table[row sep=crcr]{%
6.66166666666667	0.6549351316512\\
6.755	0.6549351316512\\
};
\addlegendentry{PSBT}

\addplot [color=mycolor2, draw=none, mark size=1.5pt, mark=+, mark options={solid, fill=white!60!black, white}]
  table[row sep=crcr]{%
0.825	0.238050501175292\\
0.825	0.268018471771687\\
0.825	0.302323673928969\\
};
\addplot [color=black, dashed]
  table[row sep=crcr]{%
0.825	0.227983997079056\\
0.825	0.0444049129728228\\
};
\addplot [color=black]
  table[row sep=crcr]{%
0.801666666666667	0.227983997079056\\
0.848333333333333	0.227983997079056\\
};
\addplot [color=black]
  table[row sep=crcr]{%
0.801666666666667	0.0444049129728228\\
0.848333333333333	0.0444049129728228\\
};
\draw[fill=mycolor3, draw=black] (axis cs:0.778333333333333,0.0595525018870831) rectangle (axis cs:0.871666666666667,0.127284371294081);
\addplot [color=black, line width=1.0pt]
  table[row sep=crcr]{%
0.778333333333333	0.0996636833879165\\
0.871666666666667	0.0996636833879165\\
};
\addplot [color=mycolor4, draw=none, mark size=1.5pt, mark=+, mark options={solid, fill=white!60!black, white}]
  table[row sep=crcr]{%
1.825	0.420882566076627\\
1.825	0.436708674895077\\
1.825	0.482459070767981\\
1.825	0.532333572990091\\
};
\addplot [color=black, dashed]
  table[row sep=crcr]{%
1.825	0.385643246112866\\
1.825	0.0858445139601827\\
};
\addplot [color=black]
  table[row sep=crcr]{%
1.80166666666667	0.385643246112866\\
1.84833333333333	0.385643246112866\\
};
\addplot [color=black]
  table[row sep=crcr]{%
1.80166666666667	0.0858445139601827\\
1.84833333333333	0.0858445139601827\\
};
\draw[fill=mycolor3, draw=black] (axis cs:1.77833333333333,0.117317352443933) rectangle (axis cs:1.87166666666667,0.234115045517683);
\addplot [color=black, line width=1.0pt]
  table[row sep=crcr]{%
1.77833333333333	0.188641558401287\\
1.87166666666667	0.188641558401287\\
};
\addplot [color=mycolor5, draw=none, mark size=1.5pt, mark=+, mark options={solid, fill=white!60!black, white}]
  table[row sep=crcr]{%
2.825	0.5655828363595\\
2.825	0.583571985508378\\
2.825	0.634077359740699\\
2.825	0.686513819117395\\
};
\addplot [color=black, dashed]
  table[row sep=crcr]{%
2.825	0.524595730667151\\
2.825	0.127628164365888\\
};
\addplot [color=black]
  table[row sep=crcr]{%
2.80166666666667	0.524595730667151\\
2.84833333333333	0.524595730667151\\
};
\addplot [color=black]
  table[row sep=crcr]{%
2.80166666666667	0.127628164365888\\
2.84833333333333	0.127628164365888\\
};
\draw[fill=mycolor3, draw=black] (axis cs:2.77833333333333,0.171758193522692) rectangle (axis cs:2.87166666666667,0.326670490205288);
\addplot [color=black, line width=1.0pt]
  table[row sep=crcr]{%
2.77833333333333	0.270116120576859\\
2.87166666666667	0.270116120576859\\
};
\addplot [color=mycolor6, draw=none, mark size=1.5pt, mark=+, mark options={solid, fill=white!60!black, white}]
  table[row sep=crcr]{%
3.825	0.692144573287839\\
3.825	0.74127770174036\\
3.825	0.789863928799235\\
};
\addplot [color=black, dashed]
  table[row sep=crcr]{%
3.825	0.674127800320328\\
3.825	0.172234520316124\\
};
\addplot [color=black]
  table[row sep=crcr]{%
3.80166666666667	0.674127800320328\\
3.84833333333333	0.674127800320328\\
};
\addplot [color=black]
  table[row sep=crcr]{%
3.80166666666667	0.172234520316124\\
3.84833333333333	0.172234520316124\\
};
\draw[fill=mycolor3, draw=black] (axis cs:3.77833333333333,0.22351835668087) rectangle (axis cs:3.87166666666667,0.409282729029655);
\addplot [color=black, line width=1.0pt]
  table[row sep=crcr]{%
3.77833333333333	0.345775216817856\\
3.87166666666667	0.345775216817856\\
};
\addplot [color=mycolor7, draw=none, mark size=1.5pt, mark=+, mark options={solid, fill=white!60!black, white}]
  table[row sep=crcr]{%
4.825	0.817072735457645\\
4.825	0.859141579078955\\
};
\addplot [color=black, dashed]
  table[row sep=crcr]{%
4.825	0.77240973119489\\
4.825	0.215252876281738\\
};
\addplot [color=black]
  table[row sep=crcr]{%
4.80166666666667	0.77240973119489\\
4.84833333333333	0.77240973119489\\
};
\addplot [color=black]
  table[row sep=crcr]{%
4.80166666666667	0.215252876281738\\
4.84833333333333	0.215252876281738\\
};
\draw[fill=mycolor3, draw=black] (axis cs:4.77833333333333,0.273548163473606) rectangle (axis cs:4.87166666666667,0.483774930238724);
\addplot [color=black, line width=1.0pt]
  table[row sep=crcr]{%
4.77833333333333	0.415501810610294\\
4.87166666666667	0.415501810610294\\
};
\addplot [color=mycolor8, draw=none, mark size=1.5pt, mark=+, mark options={solid, fill=white!60!black, white}]
  table[row sep=crcr]{%
5.825	0.905579776803697\\
};
\addplot [color=black, dashed]
  table[row sep=crcr]{%
5.825	0.870662929565632\\
5.825	0.256110608577728\\
};
\addplot [color=black]
  table[row sep=crcr]{%
5.80166666666667	0.870662929565632\\
5.84833333333333	0.870662929565632\\
};
\addplot [color=black]
  table[row sep=crcr]{%
5.80166666666667	0.256110608577728\\
5.84833333333333	0.256110608577728\\
};
\draw[fill=mycolor3, draw=black] (axis cs:5.77833333333333,0.321995049715042) rectangle (axis cs:5.87166666666667,0.550764352083206);
\addplot [color=black, line width=1.0pt]
  table[row sep=crcr]{%
5.77833333333333	0.478809259831905\\
5.87166666666667	0.478809259831905\\
};
\addplot [color=black, dashed]
  table[row sep=crcr]{%
6.825	0.93670823164036\\
6.825	0.295925959944725\\
};
\addplot [color=black]
  table[row sep=crcr]{%
6.80166666666667	0.93670823164036\\
6.84833333333333	0.93670823164036\\
};
\addplot [color=black]
  table[row sep=crcr]{%
6.80166666666667	0.295925959944725\\
6.84833333333333	0.295925959944725\\
};
\draw[fill=mycolor3, draw=black] (axis cs:6.77833333333333,0.36742477118969) rectangle (axis cs:6.87166666666667,0.60840705037117);
\addplot [color=black, line width=1.0pt]
  table[row sep=crcr]{%
6.77833333333333	0.537667408585548\\
6.87166666666667	0.537667408585548\\
};
\addlegendentry{GC}

\addplot [color=black, dashed]
  table[row sep=crcr]{%
0.941666666666667	0.177696002647281\\
0.941666666666667	0.0710701728239655\\
};
\addplot [color=black]
  table[row sep=crcr]{%
0.918333333333333	0.177696002647281\\
0.965	0.177696002647281\\
};
\addplot [color=black]
  table[row sep=crcr]{%
0.918333333333333	0.0710701728239655\\
0.965	0.0710701728239655\\
};
\draw[fill=mycolor9, draw=black] (axis cs:0.895,0.102459711022675) rectangle (axis cs:0.988333333333333,0.14837622421328);
\addplot [color=black, line width=1.0pt]
  table[row sep=crcr]{%
0.895	0.131589061522391\\
0.988333333333333	0.131589061522391\\
};
\addplot [color=black, dashed]
  table[row sep=crcr]{%
1.94166666666667	0.32200913131237\\
1.94166666666667	0.141381647437811\\
};
\addplot [color=black]
  table[row sep=crcr]{%
1.91833333333333	0.32200913131237\\
1.965	0.32200913131237\\
};
\addplot [color=black]
  table[row sep=crcr]{%
1.91833333333333	0.141381647437811\\
1.965	0.141381647437811\\
};
\draw[fill=mycolor9, draw=black] (axis cs:1.895,0.187689017504454) rectangle (axis cs:1.98833333333333,0.269986959639937);
\addplot [color=black, line width=1.0pt]
  table[row sep=crcr]{%
1.895	0.243417623220012\\
1.98833333333333	0.243417623220012\\
};
\addplot [color=black, dashed]
  table[row sep=crcr]{%
2.94166666666667	0.440707460045815\\
2.94166666666667	0.215634405612946\\
};
\addplot [color=black]
  table[row sep=crcr]{%
2.91833333333333	0.440707460045815\\
2.965	0.440707460045815\\
};
\addplot [color=black]
  table[row sep=crcr]{%
2.91833333333333	0.215634405612946\\
2.965	0.215634405612946\\
};
\draw[fill=mycolor9, draw=black] (axis cs:2.895,0.286868549883366) rectangle (axis cs:2.98833333333333,0.374436527490616);
\addplot [color=black, line width=1.0pt]
  table[row sep=crcr]{%
2.895	0.341540310531855\\
2.98833333333333	0.341540310531855\\
};
\addplot [color=black, dashed]
  table[row sep=crcr]{%
3.94166666666667	0.539080634713173\\
3.94166666666667	0.292743943631649\\
};
\addplot [color=black]
  table[row sep=crcr]{%
3.91833333333333	0.539080634713173\\
3.965	0.539080634713173\\
};
\addplot [color=black]
  table[row sep=crcr]{%
3.91833333333333	0.292743943631649\\
3.965	0.292743943631649\\
};
\draw[fill=mycolor9, draw=black] (axis cs:3.895,0.375755295157433) rectangle (axis cs:3.98833333333333,0.46612960100174);
\addplot [color=black, line width=1.0pt]
  table[row sep=crcr]{%
3.895	0.426655007526278\\
3.98833333333333	0.426655007526278\\
};
\addplot [color=black, dashed]
  table[row sep=crcr]{%
4.94166666666667	0.620688661932945\\
4.94166666666667	0.368128806352615\\
};
\addplot [color=black]
  table[row sep=crcr]{%
4.91833333333333	0.620688661932945\\
4.965	0.620688661932945\\
};
\addplot [color=black]
  table[row sep=crcr]{%
4.91833333333333	0.368128806352615\\
4.965	0.368128806352615\\
};
\draw[fill=mycolor9, draw=black] (axis cs:4.895,0.455988690257072) rectangle (axis cs:4.98833333333333,0.542866554111242);
\addplot [color=black, line width=1.0pt]
  table[row sep=crcr]{%
4.895	0.502242738381028\\
4.98833333333333	0.502242738381028\\
};
\addplot [color=black, dashed]
  table[row sep=crcr]{%
5.94166666666667	0.682602941989899\\
5.94166666666667	0.433084905147552\\
};
\addplot [color=black]
  table[row sep=crcr]{%
5.91833333333333	0.682602941989899\\
5.965	0.682602941989899\\
};
\addplot [color=black]
  table[row sep=crcr]{%
5.91833333333333	0.433084905147552\\
5.965	0.433084905147552\\
};
\draw[fill=mycolor9, draw=black] (axis cs:5.895,0.523671090602875) rectangle (axis cs:5.98833333333333,0.605259746313095);
\addplot [color=black, line width=1.0pt]
  table[row sep=crcr]{%
5.895	0.566554609686136\\
5.98833333333333	0.566554609686136\\
};
\addplot [color=black, dashed]
  table[row sep=crcr]{%
6.94166666666667	0.732859551906586\\
6.94166666666667	0.492077797651291\\
};
\addplot [color=black]
  table[row sep=crcr]{%
6.91833333333333	0.732859551906586\\
6.965	0.732859551906586\\
};
\addplot [color=black]
  table[row sep=crcr]{%
6.91833333333333	0.492077797651291\\
6.965	0.492077797651291\\
};
\draw[fill=mycolor9, draw=black] (axis cs:6.895,0.583737730979919) rectangle (axis cs:6.98833333333333,0.658421210944653);
\addplot [color=black, line width=1.0pt]
  table[row sep=crcr]{%
6.895	0.622065022587776\\
6.98833333333333	0.622065022587776\\
};
\addlegendentry{GM}

\addplot [color=black, dashed]
  table[row sep=crcr]{%
1.05833333333333	0.176887112595837\\
1.05833333333333	0.0489629828429363\\
};
\addplot [color=black]
  table[row sep=crcr]{%
1.035	0.176887112595837\\
1.08166666666667	0.176887112595837\\
};
\addplot [color=black]
  table[row sep=crcr]{%
1.035	0.0489629828429363\\
1.08166666666667	0.0489629828429363\\
};
\draw[fill=mycolor10, draw=black] (axis cs:1.01166666666667,0.0942464297012145) rectangle (axis cs:1.105,0.149886850349302);
\addplot [color=black, line width=1.0pt]
  table[row sep=crcr]{%
1.01166666666667	0.117814799212571\\
1.105	0.117814799212571\\
};
\addplot [color=black, dashed]
  table[row sep=crcr]{%
2.05833333333333	0.318141946149607\\
2.05833333333333	0.102807269850598\\
};
\addplot [color=black]
  table[row sep=crcr]{%
2.035	0.318141946149607\\
2.08166666666667	0.318141946149607\\
};
\addplot [color=black]
  table[row sep=crcr]{%
2.035	0.102807269850598\\
2.08166666666667	0.102807269850598\\
};
\draw[fill=mycolor10, draw=black] (axis cs:2.01166666666667,0.180803498850031) rectangle (axis cs:2.105,0.273557560732301);
\addplot [color=black, line width=1.0pt]
  table[row sep=crcr]{%
2.01166666666667	0.222904524551077\\
2.105	0.222904524551077\\
};
\addplot [color=black, dashed]
  table[row sep=crcr]{%
3.05833333333333	0.429967901475777\\
3.05833333333333	0.158158126003968\\
};
\addplot [color=black]
  table[row sep=crcr]{%
3.035	0.429967901475777\\
3.08166666666667	0.429967901475777\\
};
\addplot [color=black]
  table[row sep=crcr]{%
3.035	0.158158126003968\\
3.08166666666667	0.158158126003968\\
};
\draw[fill=mycolor10, draw=black] (axis cs:3.01166666666667,0.260363397852302) rectangle (axis cs:3.105,0.375919575874107);
\addplot [color=black, line width=1.0pt]
  table[row sep=crcr]{%
3.01166666666667	0.318965713222269\\
3.105	0.318965713222269\\
};
\addplot [color=black, dashed]
  table[row sep=crcr]{%
4.05833333333333	0.517091059796257\\
4.05833333333333	0.211566683408364\\
};
\addplot [color=black]
  table[row sep=crcr]{%
4.035	0.517091059796257\\
4.08166666666667	0.517091059796257\\
};
\addplot [color=black]
  table[row sep=crcr]{%
4.035	0.211566683408364\\
4.08166666666667	0.211566683408364\\
};
\draw[fill=mycolor10, draw=black] (axis cs:4.01166666666667,0.331643100857491) rectangle (axis cs:4.105,0.462884053051766);
\addplot [color=black, line width=1.0pt]
  table[row sep=crcr]{%
4.01166666666667	0.399876770879587\\
4.105	0.399876770879587\\
};
\addplot [color=black, dashed]
  table[row sep=crcr]{%
5.05833333333333	0.596010290086269\\
5.05833333333333	0.265895672982879\\
};
\addplot [color=black]
  table[row sep=crcr]{%
5.035	0.596010290086269\\
5.08166666666667	0.596010290086269\\
};
\addplot [color=black]
  table[row sep=crcr]{%
5.035	0.265895672982879\\
5.08166666666667	0.265895672982879\\
};
\draw[fill=mycolor10, draw=black] (axis cs:5.01166666666667,0.39603572148917) rectangle (axis cs:5.105,0.533583292344701);
\addplot [color=black, line width=1.0pt]
  table[row sep=crcr]{%
5.01166666666667	0.471011048843138\\
5.105	0.471011048843138\\
};
\addplot [color=black, dashed]
  table[row sep=crcr]{%
6.05833333333333	0.697089925408363\\
6.05833333333333	0.320664023318386\\
};
\addplot [color=black]
  table[row sep=crcr]{%
6.035	0.697089925408363\\
6.08166666666667	0.697089925408363\\
};
\addplot [color=black]
  table[row sep=crcr]{%
6.035	0.320664023318386\\
6.08166666666667	0.320664023318386\\
};
\draw[fill=mycolor10, draw=black] (axis cs:6.01166666666667,0.455059524058015) rectangle (axis cs:6.105,0.591238788299961);
\addplot [color=black, line width=1.0pt]
  table[row sep=crcr]{%
6.01166666666667	0.531396203570694\\
6.105	0.531396203570694\\
};
\addplot [color=black, dashed]
  table[row sep=crcr]{%
7.05833333333333	0.783646941184998\\
7.05833333333333	0.375160399067681\\
};
\addplot [color=black]
  table[row sep=crcr]{%
7.035	0.783646941184998\\
7.08166666666667	0.783646941184998\\
};
\addplot [color=black]
  table[row sep=crcr]{%
7.035	0.375160399067681\\
7.08166666666667	0.375160399067681\\
};
\draw[fill=mycolor10, draw=black] (axis cs:7.01166666666667,0.510332621895941) rectangle (axis cs:7.105,0.63886716941488);
\addplot [color=black, line width=1.0pt]
  table[row sep=crcr]{%
7.01166666666667	0.582157449120132\\
7.105	0.582157449120132\\
};
\addlegendentry{NW}

\addplot [color=black, dashed]
  table[row sep=crcr]{%
1.175	0.333023189141526\\
1.175	0.00427654371773389\\
};
\addplot [color=black]
  table[row sep=crcr]{%
1.15166666666667	0.333023189141526\\
1.19833333333333	0.333023189141526\\
};
\addplot [color=black]
  table[row sep=crcr]{%
1.15166666666667	0.00427654371773389\\
1.19833333333333	0.00427654371773389\\
};
\draw[fill=mycolor11, draw=black] (axis cs:1.12833333333333,0.0246900879716674) rectangle (axis cs:1.22166666666667,0.195472395080172);
\addplot [color=black, line width=1.0pt]
  table[row sep=crcr]{%
1.12833333333333	0.0492459524834294\\
1.22166666666667	0.0492459524834294\\
};
\addplot [color=black, dashed]
  table[row sep=crcr]{%
2.175	0.574716808917726\\
2.175	0.00900681250249491\\
};
\addplot [color=black]
  table[row sep=crcr]{%
2.15166666666667	0.574716808917726\\
2.19833333333333	0.574716808917726\\
};
\addplot [color=black]
  table[row sep=crcr]{%
2.15166666666667	0.00900681250249491\\
2.19833333333333	0.00900681250249491\\
};
\draw[fill=mycolor11, draw=black] (axis cs:2.12833333333333,0.0514092129852918) rectangle (axis cs:2.22166666666667,0.368190065531874);
\addplot [color=black, line width=1.0pt]
  table[row sep=crcr]{%
2.12833333333333	0.101124331643915\\
2.22166666666667	0.101124331643915\\
};
\addplot [color=black, dashed]
  table[row sep=crcr]{%
3.175	0.7288274649544\\
3.175	0.0137146097439422\\
};
\addplot [color=black]
  table[row sep=crcr]{%
3.15166666666667	0.7288274649544\\
3.19833333333333	0.7288274649544\\
};
\addplot [color=black]
  table[row sep=crcr]{%
3.15166666666667	0.0137146097439422\\
3.19833333333333	0.0137146097439422\\
};
\draw[fill=mycolor11, draw=black] (axis cs:3.12833333333333,0.0773963535981743) rectangle (axis cs:3.22166666666667,0.503828344917517);
\addplot [color=black, line width=1.0pt]
  table[row sep=crcr]{%
3.12833333333333	0.150171678313023\\
3.22166666666667	0.150171678313023\\
};
\addplot [color=black, dashed]
  table[row sep=crcr]{%
4.175	0.827092757708284\\
4.175	0.0184000421950499\\
};
\addplot [color=black]
  table[row sep=crcr]{%
4.15166666666667	0.827092757708284\\
4.19833333333333	0.827092757708284\\
};
\addplot [color=black]
  table[row sep=crcr]{%
4.15166666666667	0.0184000421950499\\
4.19833333333333	0.0184000421950499\\
};
\draw[fill=mycolor11, draw=black] (axis cs:4.12833333333333,0.102671562905716) rectangle (axis cs:4.22166666666667,0.610347514534514);
\addplot [color=black, line width=1.0pt]
  table[row sep=crcr]{%
4.12833333333333	0.196542497845867\\
4.22166666666667	0.196542497845867\\
};
\addplot [color=black, dashed]
  table[row sep=crcr]{%
5.175	0.889749474695515\\
5.175	0.0230632161016526\\
};
\addplot [color=black]
  table[row sep=crcr]{%
5.15166666666667	0.889749474695515\\
5.19833333333333	0.889749474695515\\
};
\addplot [color=black]
  table[row sep=crcr]{%
5.15166666666667	0.0230632161016526\\
5.19833333333333	0.0230632161016526\\
};
\draw[fill=mycolor11, draw=black] (axis cs:5.12833333333333,0.127254344638284) rectangle (axis cs:5.22166666666667,0.693998925827009);
\addplot [color=black, line width=1.0pt]
  table[row sep=crcr]{%
5.12833333333333	0.240382862591177\\
5.22166666666667	0.240382862591177\\
};
\addplot [color=black, dashed]
  table[row sep=crcr]{%
6.175	0.929701161334772\\
6.175	0.0277042372048548\\
};
\addplot [color=black]
  table[row sep=crcr]{%
6.15166666666667	0.929701161334772\\
6.19833333333333	0.929701161334772\\
};
\addplot [color=black]
  table[row sep=crcr]{%
6.15166666666667	0.0277042372048548\\
6.19833333333333	0.0277042372048548\\
};
\draw[fill=mycolor11, draw=black] (axis cs:6.12833333333333,0.151163668211353) rectangle (axis cs:6.22166666666667,0.759691877024307);
\addplot [color=black, line width=1.0pt]
  table[row sep=crcr]{%
6.12833333333333	0.281830872213058\\
6.22166666666667	0.281830872213058\\
};
\addplot [color=black, dashed]
  table[row sep=crcr]{%
7.175	0.955175481440733\\
7.175	0.0323232107434288\\
};
\addplot [color=black]
  table[row sep=crcr]{%
7.15166666666667	0.955175481440733\\
7.19833333333333	0.955175481440733\\
};
\addplot [color=black]
  table[row sep=crcr]{%
7.15166666666667	0.0323232107434288\\
7.19833333333333	0.0323232107434288\\
};
\draw[fill=mycolor11, draw=black] (axis cs:7.12833333333333,0.174417983363344) rectangle (axis cs:7.22166666666667,0.811281727934543);
\addplot [color=black, line width=1.0pt]
  table[row sep=crcr]{%
7.12833333333333	0.321017088883432\\
7.22166666666667	0.321017088883432\\
};
\addlegendentry{W}

\addplot [color=black, dashed]
  table[row sep=crcr]{%
1.29166666666667	0.174840339527009\\
1.29166666666667	0.00592935848817433\\
};
\addplot [color=black]
  table[row sep=crcr]{%
1.26833333333333	0.174840339527009\\
1.315	0.174840339527009\\
};
\addplot [color=black]
  table[row sep=crcr]{%
1.26833333333333	0.00592935848817433\\
1.315	0.00592935848817433\\
};
\draw[fill=white, draw=black] (axis cs:1.245,0.0205147844535531) rectangle (axis cs:1.33833333333333,0.118516876199181);
\addplot [color=black, line width=1.0pt]
  table[row sep=crcr]{%
1.245	0.0252290071148309\\
1.33833333333333	0.0252290071148309\\
};
\addplot [color=black, dashed]
  table[row sep=crcr]{%
2.29166666666667	0.305825283872764\\
2.29166666666667	0.0130595310438366\\
};
\addplot [color=black]
  table[row sep=crcr]{%
2.26833333333333	0.305825283872764\\
2.315	0.305825283872764\\
};
\addplot [color=black]
  table[row sep=crcr]{%
2.26833333333333	0.0130595310438366\\
2.315	0.0130595310438366\\
};
\draw[fill=white, draw=black] (axis cs:2.245,0.0391057473607361) rectangle (axis cs:2.33833333333333,0.220360899344087);
\addplot [color=black, line width=1.0pt]
  table[row sep=crcr]{%
2.245	0.0532651324210747\\
2.33833333333333	0.0532651324210747\\
};
\addplot [color=black, dashed]
  table[row sep=crcr]{%
3.29166666666667	0.408085200935602\\
3.29166666666667	0.0209190765926905\\
};
\addplot [color=black]
  table[row sep=crcr]{%
3.26833333333333	0.408085200935602\\
3.315	0.408085200935602\\
};
\addplot [color=black]
  table[row sep=crcr]{%
3.26833333333333	0.0209190765926905\\
3.315	0.0209190765926905\\
};
\draw[fill=white, draw=black] (axis cs:3.245,0.0555704385042191) rectangle (axis cs:3.33833333333333,0.300488598644733);
\addplot [color=black, line width=1.0pt]
  table[row sep=crcr]{%
3.245	0.0832228621584363\\
3.33833333333333	0.0832228621584363\\
};
\addplot [color=black, dashed]
  table[row sep=crcr]{%
4.29166666666667	0.501552075147629\\
4.29166666666667	0.0275979746511439\\
};
\addplot [color=black]
  table[row sep=crcr]{%
4.26833333333333	0.501552075147629\\
4.315	0.501552075147629\\
};
\addplot [color=black]
  table[row sep=crcr]{%
4.26833333333333	0.0275979746511439\\
4.315	0.0275979746511439\\
};
\draw[fill=white, draw=black] (axis cs:4.245,0.0700447517447174) rectangle (axis cs:4.33833333333333,0.35835421432057);
\addplot [color=black, line width=1.0pt]
  table[row sep=crcr]{%
4.245	0.106478899833746\\
4.33833333333333	0.106478899833746\\
};
\addplot [color=black, dashed]
  table[row sep=crcr]{%
5.29166666666667	0.581205271184444\\
5.29166666666667	0.0341711260844022\\
};
\addplot [color=black]
  table[row sep=crcr]{%
5.26833333333333	0.581205271184444\\
5.315	0.581205271184444\\
};
\addplot [color=black]
  table[row sep=crcr]{%
5.26833333333333	0.0341711260844022\\
5.315	0.0341711260844022\\
};
\draw[fill=white, draw=black] (axis cs:5.245,0.0835029627196491) rectangle (axis cs:5.33833333333333,0.417412422597408);
\addplot [color=black, line width=1.0pt]
  table[row sep=crcr]{%
5.245	0.134618879528716\\
5.33833333333333	0.134618879528716\\
};
\addplot [color=black, dashed]
  table[row sep=crcr]{%
6.29166666666667	0.645625737106535\\
6.29166666666667	0.0406628575874493\\
};
\addplot [color=black]
  table[row sep=crcr]{%
6.26833333333333	0.645625737106535\\
6.315	0.645625737106535\\
};
\addplot [color=black]
  table[row sep=crcr]{%
6.26833333333333	0.0406628575874493\\
6.315	0.0406628575874493\\
};
\draw[fill=white, draw=black] (axis cs:6.245,0.0994358733296394) rectangle (axis cs:6.33833333333333,0.460762079232154);
\addplot [color=black, line width=1.0pt]
  table[row sep=crcr]{%
6.245	0.163524949923158\\
6.33833333333333	0.163524949923158\\
};
\addplot [color=black, dashed]
  table[row sep=crcr]{%
7.29166666666667	0.699976455990509\\
7.29166666666667	0.0468007484450936\\
};
\addplot [color=black]
  table[row sep=crcr]{%
7.26833333333333	0.699976455990509\\
7.315	0.699976455990509\\
};
\addplot [color=black]
  table[row sep=crcr]{%
7.26833333333333	0.0468007484450936\\
7.315	0.0468007484450936\\
};
\draw[fill=white, draw=black] (axis cs:7.245,0.113357642665505) rectangle (axis cs:7.33833333333333,0.493951919488609);
\addplot [color=black, line width=1.0pt]
  table[row sep=crcr]{%
7.245	0.193158401176333\\
7.33833333333333	0.193158401176333\\
};
\addlegendentry{RND}

\addplot [color=white!1!black]
  table[row sep=crcr]{%
1.5	0\\
1.5	1\\
};
\addplot [color=white!1!black]
  table[row sep=crcr]{%
2.5	0\\
2.5	1\\
};
\addplot [color=white!1!black]
  table[row sep=crcr]{%
3.5	0\\
3.5	1\\
};
\addplot [color=white!1!black]
  table[row sep=crcr]{%
4.5	0\\
4.5	1\\
};
\addplot [color=white!1!black]
  table[row sep=crcr]{%
5.5	0\\
5.5	1\\
};
\addplot [color=white!1!black]
  table[row sep=crcr]{%
6.5	0\\
6.5	1\\
};
\end{axis}
\end{tikzpicture}%

%% file: fig/tikz/t150_mtts.tex
%
%
\definecolor{mycolor1}{rgb}{0.78431,0.82745,0.09020}%
\definecolor{mycolor2}{rgb}{0.78431,0.15686,0.12549}%
\definecolor{mycolor3}{rgb}{0.90588,0.48235,0.16078}%
\definecolor{mycolor4}{rgb}{0.00000,0.31373,0.60784}%
\begin{tikzpicture}

\begin{axis}[%
width=\figurewidth,
height=0.993\figureheight,
at={(0\figurewidth,0\figureheight)},
scale only axis,
xmin=0.5,
xmax=4.5,
xtick={1,2,3,4},
xticklabels={{PSBT},{GC},{GM},{NW}},
ymin=0,
ymax=450,
ylabel style={font=\color{white!15!black}},
ylabel={$\mu_{\mathrm{T}}^{-1}$/\SI{}{\second}},
axis background/.style={fill=white},
ylabel style={rotate=-90, at={(axis description cs:0.01,0.8),font=\scriptsize}, anchor=north east,inner ysep=0pt}
]
\addplot [color=black, dashed, forget plot]
  table[row sep=crcr]{%
1	234.470682984818\\
1	316.979882313192\\
};
\addplot [color=black, dashed, forget plot]
  table[row sep=crcr]{%
2	285.733917828848\\
2	398.822101014637\\
};
\addplot [color=black, dashed, forget plot]
  table[row sep=crcr]{%
3	218.063030687992\\
3	289.302378381853\\
};
\addplot [color=black, dashed, forget plot]
  table[row sep=crcr]{%
4	272.112661262804\\
4	342.863903577981\\
};
\addplot [color=black, dashed, forget plot]
  table[row sep=crcr]{%
1	102.33711524418\\
1	160.179832556108\\
};
\addplot [color=black, dashed, forget plot]
  table[row sep=crcr]{%
2	44.4444444444445\\
2	143.614079122485\\
};
\addplot [color=black, dashed, forget plot]
  table[row sep=crcr]{%
3	107.110081391126\\
3	152.303173315749\\
};
\addplot [color=black, dashed, forget plot]
  table[row sep=crcr]{%
4	191.377736440271\\
4	224.151111061619\\
};
\addplot [color=black, forget plot]
  table[row sep=crcr]{%
0.875	316.979882313192\\
1.125	316.979882313192\\
};
\addplot [color=black, forget plot]
  table[row sep=crcr]{%
1.875	398.822101014637\\
2.125	398.822101014637\\
};
\addplot [color=black, forget plot]
  table[row sep=crcr]{%
2.875	289.302378381853\\
3.125	289.302378381853\\
};
\addplot [color=black, forget plot]
  table[row sep=crcr]{%
3.875	342.863903577981\\
4.125	342.863903577981\\
};
\addplot [color=black, forget plot]
  table[row sep=crcr]{%
0.875	102.33711524418\\
1.125	102.33711524418\\
};
\addplot [color=black, forget plot]
  table[row sep=crcr]{%
1.875	44.4444444444444\\
2.125	44.4444444444444\\
};
\addplot [color=black, forget plot]
  table[row sep=crcr]{%
2.875	107.110081391126\\
3.125	107.110081391126\\
};
\addplot [color=black, forget plot]
  table[row sep=crcr]{%
3.875	191.377736440271\\
4.125	191.377736440271\\
};
\addplot [color=black, forget plot]
  table[row sep=crcr]{%
0.75	160.179832556108\\
0.75	234.470682984818\\
1.25	234.470682984818\\
1.25	160.179832556108\\
0.75	160.179832556108\\
};
\addplot [color=black, forget plot]
  table[row sep=crcr]{%
1.75	143.614079122485\\
1.75	285.733917828848\\
2.25	285.733917828848\\
2.25	143.614079122485\\
1.75	143.614079122485\\
};
\addplot [color=black, forget plot]
  table[row sep=crcr]{%
2.75	152.303173315749\\
2.75	218.063030687992\\
3.25	218.063030687992\\
3.25	152.303173315749\\
2.75	152.303173315749\\
};
\addplot [color=black, forget plot]
  table[row sep=crcr]{%
3.75	224.151111061619\\
3.75	272.112661262804\\
4.25	272.112661262804\\
4.25	224.151111061619\\
3.75	224.151111061619\\
};
\addplot [color=black, forget plot]
  table[row sep=crcr]{%
0.75	190.558766316749\\
1.25	190.558766316749\\
};
\addplot [color=black, forget plot]
  table[row sep=crcr]{%
1.75	185.206577731936\\
2.25	185.206577731936\\
};
\addplot [color=black, forget plot]
  table[row sep=crcr]{%
2.75	193.048512066679\\
3.25	193.048512066679\\
};
\addplot [color=black, forget plot]
  table[row sep=crcr]{%
3.75	244.735249833522\\
4.25	244.735249833522\\
};
\addplot [color=black, draw=none, mark=+, mark options={solid, red}, forget plot]
  table[row sep=crcr]{%
1	358.844430076994\\
1	387.803805135839\\
1	387.803805135839\\
};
\addplot [color=black, draw=none, mark=+, mark options={solid, red}, forget plot]
  table[row sep=crcr]{%
4	344.171595302719\\
4	346.781124558009\\
4	347.797349946496\\
4	348.657136552071\\
4	354.705815419355\\
4	359.978922955797\\
4	360.789811171601\\
4	371.123923098656\\
};

\addplot[area legend, draw=black, fill=mycolor1, forget plot]
table[row sep=crcr] {%
x	y\\
3.75	224.151111061619\\
3.75	272.112661262804\\
4.25	272.112661262804\\
4.25	224.151111061619\\
3.75	224.151111061619\\
}--cycle;

\addplot[area legend, draw=black, fill=mycolor2, forget plot]
table[row sep=crcr] {%
x	y\\
2.75	152.303173315749\\
2.75	218.063030687992\\
3.25	218.063030687992\\
3.25	152.303173315749\\
2.75	152.303173315749\\
}--cycle;

\addplot[area legend, draw=black, fill=mycolor3, forget plot]
table[row sep=crcr] {%
x	y\\
1.75	143.614079122485\\
1.75	285.733917828848\\
2.25	285.733917828848\\
2.25	143.614079122485\\
1.75	143.614079122485\\
}--cycle;

\addplot[area legend, draw=black, fill=mycolor4, forget plot]
table[row sep=crcr] {%
x	y\\
0.75	160.179832556108\\
0.75	234.470682984818\\
1.25	234.470682984818\\
1.25	160.179832556108\\
0.75	160.179832556108\\
}--cycle;

\addplot[area legend, draw=black, fill=black, forget plot]
table[row sep=crcr] {%
x	y\\
3.75	244.735249833522\\
4.25	244.735249833522\\
}--cycle;

\addplot[area legend, draw=black, fill=black, forget plot]
table[row sep=crcr] {%
x	y\\
2.75	193.048512066679\\
3.25	193.048512066679\\
}--cycle;

\addplot[area legend, draw=black, fill=black, forget plot]
table[row sep=crcr] {%
x	y\\
1.75	185.206577731936\\
2.25	185.206577731936\\
}--cycle;

\addplot[area legend, draw=black, fill=black, forget plot]
table[row sep=crcr] {%
x	y\\
0.75	190.558766316749\\
1.25	190.558766316749\\
}--cycle;
\end{axis}
\end{tikzpicture}%

%% file: fig/tikz/02_13_sbt_drive.tex
%
%
\definecolor{mycolor1}{rgb}{0.90588,0.96863,1.00000}%
\definecolor{mycolor2}{rgb}{0.78431,0.82745,0.09020}%
\definecolor{mycolor3}{rgb}{0.78431,0.15686,0.12549}%
\definecolor{mycolor4}{rgb}{0.90588,0.48235,0.16078}%
\definecolor{mycolor5}{rgb}{0.00000,0.31373,0.60784}%
\begin{tikzpicture}

\begin{axis}[%
width=0.951\figurewidth,
height=\figureheight,
at={(0\figurewidth,0\figureheight)},
scale only axis,
xmin=0.5,
xmax=5.5,
xtick={1,2,3,4,5},
xticklabels={{PSBT},{GC},{GM},{NW},{W}},
ymin=0,
ymax=750,
ylabel style={font=\color{white!15!black}},
ylabel={$t_{\mathrm{r}}$/\SI{}{\second}},
axis background/.style={fill=white},
ylabel style={rotate=-90, at={(axis description cs:-0.0,1)}, anchor=north east,inner ysep=0pt,font=\scriptsize}
]
\addplot [color=black, dashed, forget plot]
  table[row sep=crcr]{%
1	150.46725\\
1	247.027\\
};
\addplot [color=black, dashed, forget plot]
  table[row sep=crcr]{%
2	119.3965\\
2	189.462\\
};
\addplot [color=black, dashed, forget plot]
  table[row sep=crcr]{%
3	111.66\\
3	131.583\\
};
\addplot [color=black, dashed, forget plot]
  table[row sep=crcr]{%
4	149.048\\
4	248.767\\
};
\addplot [color=black, dashed, forget plot]
  table[row sep=crcr]{%
5	514.344\\
5	825\\
};
\addplot [color=black, dashed, forget plot]
  table[row sep=crcr]{%
1	10.0554\\
1	81.071825\\
};
\addplot [color=black, dashed, forget plot]
  table[row sep=crcr]{%
2	10.3738\\
2	59.08715\\
};
\addplot [color=black, dashed, forget plot]
  table[row sep=crcr]{%
3	58.5247\\
3	89.2959\\
};
\addplot [color=black, dashed, forget plot]
  table[row sep=crcr]{%
4	35.2756\\
4	80.1827\\
};
\addplot [color=black, dashed, forget plot]
  table[row sep=crcr]{%
5	16.4999442100525\\
5	84.3655014662743\\
};
\addplot [color=black, forget plot]
  table[row sep=crcr]{%
0.875	247.027\\
1.125	247.027\\
};
\addplot [color=black, forget plot]
  table[row sep=crcr]{%
1.875	189.462\\
2.125	189.462\\
};
\addplot [color=black, forget plot]
  table[row sep=crcr]{%
2.875	131.583\\
3.125	131.583\\
};
\addplot [color=black, forget plot]
  table[row sep=crcr]{%
3.875	248.767\\
4.125	248.767\\
};
\addplot [color=black, forget plot]
  table[row sep=crcr]{%
0.875	10.0554\\
1.125	10.0554\\
};
\addplot [color=black, forget plot]
  table[row sep=crcr]{%
1.875	10.3738\\
2.125	10.3738\\
};
\addplot [color=black, forget plot]
  table[row sep=crcr]{%
2.875	58.5247\\
3.125	58.5247\\
};
\addplot [color=black, forget plot]
  table[row sep=crcr]{%
3.875	35.2756\\
4.125	35.2756\\
};
\addplot [color=black, forget plot]
  table[row sep=crcr]{%
4.875	16.4999442100525\\
5.125	16.4999442100525\\
};
\addplot [color=blue, forget plot]
  table[row sep=crcr]{%
0.75	81.071825\\
0.75	150.46725\\
1.25	150.46725\\
1.25	81.071825\\
0.75	81.071825\\
};
\addplot [color=blue, forget plot]
  table[row sep=crcr]{%
1.75	59.08715\\
1.75	119.3965\\
2.25	119.3965\\
2.25	59.08715\\
1.75	59.08715\\
};
\addplot [color=blue, forget plot]
  table[row sep=crcr]{%
2.75	89.2959\\
2.75	111.66\\
3.25	111.66\\
3.25	89.2959\\
2.75	89.2959\\
};
\addplot [color=blue, forget plot]
  table[row sep=crcr]{%
3.75	80.1827\\
3.75	149.048\\
4.25	149.048\\
4.25	80.1827\\
3.75	80.1827\\
};
\addplot [color=blue, forget plot]
  table[row sep=crcr]{%
4.75	84.3655014662743\\
4.75	514.344\\
5.25	514.344\\
5.25	84.3655014662743\\
4.75	84.3655014662743\\
};
\addplot [color=red, forget plot]
  table[row sep=crcr]{%
0.75	101.873\\
1.25	101.873\\
};
\addplot [color=red, forget plot]
  table[row sep=crcr]{%
1.75	73.3419\\
2.25	73.3419\\
};
\addplot [color=red, forget plot]
  table[row sep=crcr]{%
2.75	99.15955\\
3.25	99.15955\\
};
\addplot [color=red, forget plot]
  table[row sep=crcr]{%
3.75	98.7604\\
4.25	98.7604\\
};
\addplot [color=red, forget plot]
  table[row sep=crcr]{%
4.75	246.738\\
5.25	246.738\\
};
\addplot [color=black, draw=none, mark=+, mark options={solid, red}, forget plot]
  table[row sep=crcr]{%
1	263.325\\
1	301.385\\
};
\addplot [color=black, draw=none, mark=+, mark options={solid, red}, forget plot]
  table[row sep=crcr]{%
2	216.131\\
};
\addplot [color=black, draw=none, mark=+, mark options={solid, red}, forget plot]
  table[row sep=crcr]{%
3	35.2491\\
3	40.2748\\
3	153.319\\
3	156.969\\
3	161.125\\
};
\addplot [color=black, draw=none, mark=+, mark options={solid, red}, forget plot]
  table[row sep=crcr]{%
4	282.802\\
};

\addplot[area legend, draw=black, fill=mycolor1, forget plot]
table[row sep=crcr] {%
x	y\\
4.75	84.3655014662743\\
4.75	514.344\\
5.25	514.344\\
5.25	84.3655014662743\\
4.75	84.3655014662743\\
}--cycle;

\addplot[area legend, draw=black, fill=mycolor2, forget plot]
table[row sep=crcr] {%
x	y\\
3.75	80.1827\\
3.75	149.048\\
4.25	149.048\\
4.25	80.1827\\
3.75	80.1827\\
}--cycle;

\addplot[area legend, draw=black, fill=mycolor3, forget plot]
table[row sep=crcr] {%
x	y\\
2.75	89.2959\\
2.75	111.66\\
3.25	111.66\\
3.25	89.2959\\
2.75	89.2959\\
}--cycle;

\addplot[area legend, draw=black, fill=mycolor4, forget plot]
table[row sep=crcr] {%
x	y\\
1.75	59.08715\\
1.75	119.3965\\
2.25	119.3965\\
2.25	59.08715\\
1.75	59.08715\\
}--cycle;

\addplot[area legend, draw=black, fill=mycolor5, forget plot]
table[row sep=crcr] {%
x	y\\
0.75	81.071825\\
0.75	150.46725\\
1.25	150.46725\\
1.25	81.071825\\
0.75	81.071825\\
}--cycle;

\addplot[area legend, draw=black, fill=black, forget plot]
table[row sep=crcr] {%
x	y\\
4.75	246.738\\
5.25	246.738\\
}--cycle;

\addplot[area legend, draw=black, fill=black, forget plot]
table[row sep=crcr] {%
x	y\\
3.75	98.7604\\
4.25	98.7604\\
}--cycle;

\addplot[area legend, draw=black, fill=black, forget plot]
table[row sep=crcr] {%
x	y\\
2.75	99.15955\\
3.25	99.15955\\
}--cycle;

\addplot[area legend, draw=black, fill=black, forget plot]
table[row sep=crcr] {%
x	y\\
1.75	73.3419\\
2.25	73.3419\\
}--cycle;

\addplot[area legend, draw=black, fill=black, forget plot]
table[row sep=crcr] {%
x	y\\
0.75	101.873\\
1.25	101.873\\
}--cycle;
\end{axis}
\end{tikzpicture}%

%% file: fig/tikz/02_14_sbt_drive.tex
%
%
\definecolor{mycolor1}{rgb}{0.78431,0.82745,0.09020}%
\definecolor{mycolor2}{rgb}{0.00000,0.31373,0.60784}%
\begin{tikzpicture}

\begin{axis}[%
width=0.951\figurewidth,
height=\figureheight,
at={(0\figurewidth,0\figureheight)},
scale only axis,
xmin=0.5,
xmax=2.5,
xtick={1,2},
xticklabels={{PSBT},{NW}},
ymin=0,
ymax=250,
ylabel style={font=\color{white!15!black}},
ylabel={$t_{\mathrm{r}}$/\SI{}{\second}},
axis background/.style={fill=white},
ylabel style={rotate=-90, at={(axis description cs:-0.0,1)}, anchor=north east,inner ysep=0pt,font=\scriptsize}
]
\addplot [color=black, dashed, forget plot]
  table[row sep=crcr]{%
1	43.1394\\
1	84.6578\\
};
\addplot [color=black, dashed, forget plot]
  table[row sep=crcr]{%
2	85.170625\\
2	171.817\\
};
\addplot [color=black, dashed, forget plot]
  table[row sep=crcr]{%
1	10.0345\\
1	15.427\\
};
\addplot [color=black, dashed, forget plot]
  table[row sep=crcr]{%
2	10.0238\\
2	22.38765\\
};
\addplot [color=black, forget plot]
  table[row sep=crcr]{%
0.924999999999997	84.6578\\
1.075	84.6578\\
};
\addplot [color=black, forget plot]
  table[row sep=crcr]{%
1.92500000000001	171.817\\
2.07499999999999	171.817\\
};
\addplot [color=black, forget plot]
  table[row sep=crcr]{%
0.925000000000001	10.0345\\
1.075	10.0345\\
};
\addplot [color=black, forget plot]
  table[row sep=crcr]{%
1.925	10.0238\\
2.075	10.0238\\
};
\addplot [color=blue, forget plot]
  table[row sep=crcr]{%
0.850000000000001	15.427\\
0.850000000000001	43.1394\\
1.15	43.1394\\
1.15	15.427\\
0.850000000000001	15.427\\
};
\addplot [color=blue, forget plot]
  table[row sep=crcr]{%
1.84999999999999	22.38765\\
1.84999999999999	85.170625\\
2.15000000000001	85.170625\\
2.15000000000001	22.38765\\
1.84999999999999	22.38765\\
};
\addplot [color=red, forget plot]
  table[row sep=crcr]{%
0.850000000000001	25.2577\\
1.15	25.2577\\
};
\addplot [color=red, forget plot]
  table[row sep=crcr]{%
1.85	49.9956\\
2.15	49.9956\\
};
\addplot [color=black, draw=none, mark=+, mark options={solid, red}, forget plot]
  table[row sep=crcr]{%
1	86.2239\\
1	97.3158\\
1	98.1388\\
1	102.326\\
1	102.386\\
1	103.338\\
1	103.815\\
1	106.93\\
1	113.426\\
1	114.443\\
};
\addplot [color=black, draw=none, mark=+, mark options={solid, red}, forget plot]
  table[row sep=crcr]{%
2	180.432\\
2	182.861\\
2	197.604\\
};

\addplot[area legend, draw=black, fill=mycolor1, forget plot]
table[row sep=crcr] {%
x	y\\
1.85	22.38765\\
1.85	85.170625\\
2.15	85.170625\\
2.15	22.38765\\
1.85	22.38765\\
}--cycle;

\addplot[area legend, draw=black, fill=mycolor2, forget plot]
table[row sep=crcr] {%
x	y\\
0.85	15.427\\
0.85	43.1394\\
1.15	43.1394\\
1.15	15.427\\
0.85	15.427\\
}--cycle;

\addplot[area legend, draw=black, fill=black, forget plot]
table[row sep=crcr] {%
x	y\\
1.85	49.9956\\
2.15	49.9956\\
}--cycle;

\addplot[area legend, draw=black, fill=black, forget plot]
table[row sep=crcr] {%
x	y\\
0.85	25.2577\\
1.15	25.2577\\
}--cycle;
\end{axis}
\end{tikzpicture}%

%% file: table/places.tex
\begin{tabular}{l|lrrr|rr}
&\multicolumn{4}{c|}{\textbf{Experimental results}}&\multicolumn{2}{c}{\textbf{Model estimation}}\\ 

Place & Method & Trials &$P$& $\bar{t}_\mathrm{r}$& $\bar{\mu}_\mathrm{r}^{-1}$&$\bar{p}_{\mathrm{s,T}}(t_{\mathrm{max}})$\\ 
\hline 
 \rule{0pt}{2.5ex}$\mathcal{P}_{\mathrm{stor}}$ & PSBT & $86$ & \textbf{\SI{98.8}{\percent}} & $120.5\pm61.2$ & $139.5\pm8.7$&$0.87\pm0.01$ \\ 
 & NW & $93$ & \SI{88.2}{\percent} & $115.4\pm52.3$ & $150.8\pm7.0$&$0.85\pm0.01$ \\ 
 & GM & $86$ & \SI{62.8}{\percent} & $101.1\pm22.4$ & $121.5\pm 0.0$&$0.81\pm0.00$  \\ 
 & GC & $90$ & \SI{65.6}{\percent} & $\boldsymbol{89.1}\pm49.7$ & $108.1\pm 0.0$&$0.81\pm0.00$ \\ 
 \rule{0pt}{3.5ex}$\mathcal{P}_{\mathrm{cafe}}$ & PSBT& $112$  & \textbf{\SI{91.1}{\percent}} & $\boldsymbol{35.5}\pm28.2$&$49.9\pm0.0$&$0.99\pm0.00$ \\ 
 & NW & $121$ & \SI{90.1}{\percent} & $63.0\pm48.9$ &$126.8\pm6.0$&$0.86\pm0.01$ \\ 
\hline 

\end{tabular}